%% file: main.tex
\documentclass[12pt]{article}
\usepackage[utf8]{inputenc}
\usepackage{natbib}
\usepackage{amsmath,amssymb,amsthm}
\usepackage{pdfpages}
\usepackage{syntax}
\usepackage{graphicx}
\usepackage{enumerate}
\usepackage{varwidth}
\usepackage{varwidth}
\usepackage{newfloat}
\DeclareFloatingEnvironment[
  fileext   = logr,% don't use log here!
  listname  = {List of Grammars},
  name      = Grammar,
  placement = htp
]{Grammar}

\usepackage{algorithm}
\usepackage{algorithmicx}
\usepackage[noend]{algpseudocode}

\title{The Proof is in the Pudding: Using Automated Theorem Proving to Generate Cooking Recipes}
\author{
  Louis Mahon$^*$\\
  \footnotesize{Department of Computer Science,} \\ 
  \footnotesize{Oxford University} \\
  \and
  Carl Vogel\\
  \footnotesize{School of Computer Science and Statistics,} \\
  \footnotesize{Trinity College, Dublin}
}\date{\small * = corresponding author at louis.mahon@cs.ox.ac.uk}

\begin{document}

\maketitle

\begin{abstract}
This paper presents FASTFOOD, a rule-based Natural Language Generation Program  for cooking recipes. Recipes are generated by using an Automated Theorem  Proving procedure to select the ingredients and instructions,  with ingredients corresponding to axioms and instructions to implications. FASTFOOD also contains a temporal optimization module which can rearrange the recipe to make it more time efficient for the user, e.g. the recipe specifies to chop the vegetables while the rice is boiling. The system is described in detail, using a framework which  divides Natural Language Generation into 4 phases: content production,  content selection, content organisation and content realisation. A comparison is then made with similar existing systems and techniques. 
\end{abstract} 

\theoremstyle{definition}
\newtheorem{definition}{Definition}[section]

\theoremstyle{definition}
\newtheorem{condition}{Condition}[section]

\input{\detokenize{intro}}
\input{\detokenize{representation_system}}
\input{\detokenize{content_production}}
\input{\detokenize{content_selection}}
\input{\detokenize{content_organization}}
\input{\detokenize{content_realization}}
\input{\detokenize{worked_example.tex}}
\input{\detokenize{related_work}}
\input{\detokenize{conclusion}}

\newpage
\bibliography{bibliography}

\end{document}

%% file: intro.tex
\section{Introduction} \label{sec:intro}
In recent decades, cooking recipes have received a degree of attention in the field of rule-based Natural  Language Processing. They are highly uniform in their general structure, with an ingredients list  followed by a series of instructions, and can now be easily accessed in large numbers on cooking  websites and recipe search engines. However, the majority of work has focused either on  searching and annotating large databases of online recipes, or on methods for processing and  representing individual recipes, in either case the approach has been one of Natural Language  Understanding. The present system, named FASTFOOD, instead approaches cooking recipes  from the direction of Natural Language Generation (NLG): rather than trying to extract a  representation of the discourse of a recipe from a natural language description, it aims to derive  natural language recipes from a number of pre-existing structures. 
The task of NLG has been analysed in numerous ways. Some early examples \cite{mckeown1985discourse}) make a binary distinction between a strategic component  that decides what to say, and a tactical component that decides how to say it. Following work  has continued the idea of a spectrum from strategic to tactical but increased the number of levels  to 3 \cite{panaget1994using,bateman2002natural} or even 6 \cite{dale1991content, reiter2005choosing}. The most  fruitful framework in which to view the present work is that of \citeauthor{robin1993revision} (\citeyear{robin1993revision}), which uses 4:  content production, content selection, content organisation, and content realisation. Using this  framework, the operation of FASTFOOD can be summarised as follows. 

\textbf{Content Production} The first phase is to produce the content that may possibly appear  in discourse. FASTFOOD represents discourse content with two types of structure: descriptive  strings, intuitively a natural language description of a state of affairs; and processes, intuitively  transitions from one state of affairs to another. Internally these are produced from more  fundamental structures such as cooking actions, intuitively verbs such as chop or mash, and food  classes, intuitively a type of food such as carrot. The result of the content production phase is  the creation of a database consisting of a set containing all recognised descriptive strings and a set containing all recognised processes. 

\textbf{Content Selection} Once the possible content has been produced, the system must select  the appropriate content for a given discourse (recipe) in response to a user input. This is done by  an algorithm akin to a backwards chaining Automated Theorem Prover (ATP): the desired dish  is treated as a formula, the possible ingredients as axioms and the processes as rules of inference.  The algorithm proceeds backwards from the desired dish, attempting to reach a set of possible  ingredients. The output of the content selection phase is a set of descriptive strings corresponding to an ingredients list and a set of processes corresponding to a list of cooking  instructions; or else an indication that the dish cannot be made from the set of foods the user has access to (the formula is unprovable).  

\textbf{Content Organisation} FASTFOOD is concerned not only with generating a recipe, but  also with ensuring the recipe is time efficient. Improvements in time efficiency can be achieved  by performing multiple tasks at once: it is quicker to chop the carrots while waiting for the rice  to boil than to wait for the rice to boil and then chop the carrots. The information on how to  achieve efficiency is expressed by the third of the NLG subtasks: content organisation. FASTFOOD must determine the order in which instructions are to be performed, and provide an indication of which are to be performed simultaneously, i.e. when to print successive instructions  together in the form ‘while A, B’. The output of the content organisation phase is an ordered list  of processes, some of which have been marked as warranting concurrent execution. 

\textbf{Content Realisation} The final phase is to create text from the organised content. Each  process contains an imperative formed by an, often simple, combination of the action and food  class from which the process was formed in the content production phase. The realised text is a  successive printing of each imperative in the organised list of processes generated in the previous  phase; along with the ingredients list, the total cooking time, and the times during the cooking  at which the cook will be passive.  

Of these 4 stages, the greatest distinction is between the first and the other three. Content  production creates a database which can be stored and then used to generate various recipes in  response to user input. It need only run when this database is updated (e.g. when the program is  taught new foods, see Section 2.3). The other three stages run every time a recipe is generated.

\begin{figure}
    \centering
    \includegraphics[trim={0.5cm 0 0 0.5cm},clip, totalheight=\textheight]{\detokenize{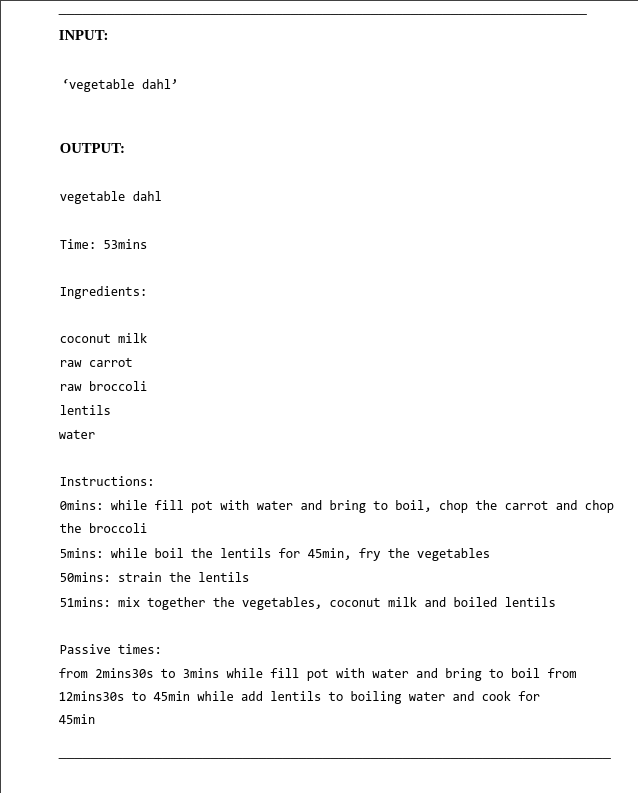}}
    \caption{An example of an input-output pair, the user inputs the title of a dish and receives a recipe for that dish  along with the total cooking time and the passive intervals that occur during the recipe. }
    \label{fig:recipe}
\end{figure}

The temporal optimisation module that constitutes the discourse organisation stage is included  for two reasons. Firstly, it demonstrates that the discourse representation system is robust enough  to allow meaningful inference of the sort that might be made by a human cook. Secondly, it is  of practical benefit to a user, and the growing area of smart kitchen appliances has specifically  called for a representation of the cooking processthat facilitates temporal optimisation \cite{hamada2005cooking, reichel2011mampf}. 
An indication of the passive times during the recipe is felt to be a useful additional piece of  information for a cook; and the temporal optimisation method already employs a notion of active  vs. passive time, so this information can be extracted naturally. Figure \ref{fig:recipe} shows an example of  a possible input-output pair for FASTFOOD.

The remainder of this paper is structured as follows.  

\textbf{Section \ref{sec:representation-system}} describes the representation system in FASTFOOD. Section \ref{subsec:discourse-ref-problems} addresses some  problems with using discourse referents in the domain of cooking recipes. Section \ref{subsec:dstrings-and-procs} introduces FASTFOOD’s representational structures, descriptive strings and processes. Section \ref{subsec:things-vs-descriptions} briefly  relates this representational system to a long standing philosophical debate on the nature of  names and descriptions.  

\textbf{Section \ref{sec:content-production}} describes how a database of all content, containing a set of descriptive strings and a set of processes, is produced. Section \ref{subsec:productive-method} gives an overview of the productive method.  Section \ref{subsec:foods-and-actions} explains how two types of structure, cooking actions and food classes interact to  produce a number of descriptive strings and processes. Section \ref{subsec:synonyms-and-ghosts} describes the elements of the  system created to deal with the existence of multiple descriptions for the same dish e.g. ‘chips’  vs. ‘fried sliced potatoes’.  

\textbf{Section \ref{sec:content-selection}} is an exposition of the algorithm used to select the discourse content for a  recipe. Section \ref{subsec:selection-algorithm} directly describes the algorithm that selects the appropriate descriptive strings and processes from FASTFOOD’s internal database in response to a user input. Section \ref{subsec:comparison-with-ATP-and-parsing}  compares this algorithm with automated methods of parsing and, in more detail, with Automated  Theorem Proving.  

\textbf{Section \ref{sec:content-organization}} describes the temporal optimisation module. Section \ref{subsec:optimization-problem-overview} gives an overview of  the optimisation problem. Sections \ref{subsec:find-all-lists} and \ref{subsec:concurrent-compression} respectively exposit the two main components of  the algorithm employed to solve this problem. 

\textbf{Section \ref{sec:content-realization}} describes the method of producing text from the organised discourse content  formed by the algorithms of the previous Sections.  

\textbf{Section \ref{sec:worked-example}} is a worked example. It walks through the steps taken by FASTFOOD in the generation of the output in Figure \ref{fig:recipe}: a recipe for the dish described by the descriptive string ‘vegetable dahl’.

\textbf{Section \ref{sec:related-work}} relates FASTFOOD to the existing literature on NLP of cooking recipes. Section  \ref{subsec:epicure-comparison} conducts a comparison with an early example of an NLG system for cooking recipes, \cite{dale1990generating}. Section \ref{subsec:current-lit} discusses an instance of the more common NLU approach,  specifically a technique for depicting recipes as workflows. A potential problem with this  technique is demonstrated, and it is shown that workflow representation is easily accomplished  in FASTFOOD.

For algorithm description, a basic style Pseudocode is used.  Syntactic definitions are, where appropriate, given in Backus-Naur Form. A Python implementation is available at https://github.com/Lou1sM/FastFood.

%% file: representation_system.tex
\section{Representation System} \label{sec:representation-system}
This section describes the format in which discourse content is represented in FASTFOOD. Section 1.1 discusses problems arising from the use of discourse referents in the domain of  cooking recipes, and explains why they are not used here. Section 1.2 exposits the structures  termed descriptive strings and processes, which together form the representation structure.  Section 1.3 sketches a brief connection between this representation choice and a long standing  philosophical debate on the nature of names, objects and descriptions.  

\subsection{Problems with Discourse Referents} \label{subsec:discourse-ref-problems}
Discourse referents constitute a popular and often useful method for semantically representing  a discourse. Each discourse referent corresponds to an entity discussed in the discourse, and the  semantics of the discourse can then be described as various changes in the properties of these  referents. Discourse referents have been used with success for a number of different purposes: 
e.g. modelling of anaphoric reduction \cite{kamp2011discourse}, topic segmentation/document summarization \cite{webber2012discourse}, and opinion mining \cite{indurkhya2010handbook, mooney1996inductive, sarawagi2008information}. However, in the domain of  cooking recipes, their use may be problematic. Cooking is essentially an activity which makes  new things out of old things, so we would expect entities to come into and out of existence over  the course of a recipe; and for a representational system based on these entities, this irregularity  in the ontology presents some difficulties.  
Firstly, how could it be determined whether a new noun phrase calls for a new referent? It does  not seem that chopped carrot should indicate something entirely new, rather it should be a  modification of the already existent, (raw) carrot; but perhaps the same cannot be said of carrot  puree or carrot soup. Some actions transform one thing into another whereas some simply  modify the original, and deriving a definite rule to distinguish is non-trivial.  
A second, related, problem is in accounting for the disappearance of individuals. After an  instruction like ‘add the salt to the boiling water’ there is no longer (at least in the immediate  sense) an individual called ‘the salt’; whereas after ‘add the quinoa to the boiling water’ the  entity corresponding to the ‘the quinoa’ is extant. The latter might be followed by ‘strain the  quinoa’, but the former could not be followed by ‘strain the salt’. Some additions constitute  combinations in which the parts are no longer easily distinguishable or separable, and some do  not; and again there is no obvious rule to capture this difference. Indeed, even among artificial intelligence programs specifically designed for chemical reasoning it is not clear that there  would be anything capable of a cook’s intuitive understanding of transformation vs.  modification, or of reversible vs. irreversible combinations \cite{mccudden2017future}. 
Despite being a fruitful approach elsewhere, the utilisation of discourse referents  encounters difficulties in the domain of cooking recipes, and so they are not employed in the  current representation.

\subsection{Descriptive Strings and Processes} \label{subsec:dstrings-and-procs}
The fundamental representational unit used in FASTFOOD is a structure called a descriptive  string. Intuitively descriptive strings can be thought of as a partial description of the state of  affairs at some stage during the carrying out of a recipe. Formally they are written as strings in  natural language—‘there is a mixture of fried vegetables and boiled lentils’, ‘there is chopped  carrot in boiling water’. However, for the sake of simplification, the initial ‘there is’ is omitted. 
Descriptive strings are then used to define processes. Intuitively, processes correspond to the  acts performed by the cook, e.g. chopping carrots. They are represented by an associative array  containing five key-value pairs. Two of the entries in this array are (input,X) and (output,Y) where X and Y are sets of  descriptive strings. Thus a process specifies a transition from one state of affairs, one set of descriptive strings, to another. Chopping carrots contains ‘raw carrot’ in the input and ‘chopped carrot’  in the output, i.e. chopping carrot is an action that begins with there being carrot and ends with  there being chopped carrot.  
It is necessary to include three additional key-value pairs to facilitate the final discourse  generation; the keys of these pairs are ‘time’, ‘f time’ (free time), and ‘direction’. The function  of these three additional elements is explained fully in the following sections—time and free  time in Section \ref{sec:content-organization} and direction in Section \ref{sec:content-realization}—but essentially they refer respectively to the time  taken to perform the process, the passive or free time that exists during this process and the  natural language imperative that describes how to perform the process. The full syntactic definition  of a process is given in (1.1). 
At times throughout this paper, processes are referred to with respect to certain elements of the array, depending on which are most relevant to the context. For example, in the content selection algorithm (Section \ref{sec:content-selection}) it is only the inputs and outputs that are relevant; whereas in the temporal optimisation algorithm (Section \ref{sec:content-organization}) it is only the time and free time that are relevant. However, semantically processes are most accurately thought of in terms of their input and output, as a  transformation from one state to another. 
\begin{grammar}
<process> ::= <input-pair> <output-pair> <time-pair> <free-time-pair> <direction-pair>

<input-pair> ::= ‘input’ <descriptive-string>

<output-pair> ::= ‘output’ <descriptive-string>

<time-pair> ::= ‘time’ integer

<free-time-pair> ::= ‘f time’ integer

<direction-pair> ::= ‘direction’ <direction>

<direction> ::= string (1.1)
\end{grammar} 
An ingredients list is represented by a set of descriptive strings, and a list of cooking instructions is represented by a list of processes. Together, the ingredients list and the cooking instructions form the discourse content selected for a given recipe in FASTFOOD (the output of the algorithm  described in Section \ref{subsec:selection-algorithm})\footnote{This content must then be organised (Section \ref{sec:content-organization}) before it is passed to the content realisation stage, Section \ref{sec:content-realization}. }.

\begin{grammar} \label{gram:selected-content}
<selected-content> ::= <ingredients-list> <action-list> 

<ingredients-list> ::= <descriptive-string>| <descriptive-string><ingredients-list> 

<action-list> ::= <process> | <process> <action-list> (1.2)
\end{grammar}
The only constraint on descriptive strings is a semantic one, that they describe a state of affairs.  In FASTFOOD this description takes place through English, but there are many other systems for  describing a state of affairs—a different natural language, logical notation\footnote{ Replacing descriptive strings with a logical notation is discussed in the comparison with EPICURE in Section \ref{subsec:epicure-comparison}. It might seem that replacing descriptive strings with statements of predicate logic such as  ‘Chopped (carrot) and chopped (broccoli)’ would mean using discourse referents (‘carrot’ and ‘broccoli’)  after all. However, as described above, the procedural component of recipe discourse is represented by a  list of processes, not by a sequence of states of affairs; so even if descriptive strings were to be replaced  with predicate logic statements, it would still be inaccurate to say that the discourse is represented in terms  of discourse referents. }, digital images or  data representing cooking related measurements e.g. the temperature of the oven.

The possibility of replacing natural language descriptions with images in FASTFOOD suggests a  compatibility with the recent attempts to examine the relationship of NLP and computer vision  in the domain of cooking recipes \cite{oh2015text, hamada2004multimedia, hamada2000structural}; and the potential to use measurement  data would possibly facilitate application to the growing industrial area of smart kitchen  appliances, \cite{hamada2005cooking, reichel2011mampf, zhaotraining}. However any  more than a suggestion of these possible extensions is outside the scope of this paper.  
In short, any structure capable of describing states of affairs could be used, and a modification  of FASTFOOD such as that briefly discussed in Section \ref{sec:related-work} could involve replacing descriptive  strings with another such structure. However, in the current version, descriptive strings are the  most convenient choice.

\subsection{Things vs Descriptions} \label{subsec:things-vs-descriptions}
A representation system in terms of states of affairs can be seen to be related to a long-standing  philosophical debate on the logical and semantic properties of names, objects and descriptions.  Introduced in its current form by Gottlob Frege (\citeyear{frege1984gottlob}), this debate is now well over 100 years  old \cite{russell1905denoting, russell1910knowledge, searle1958proper, quine1948there, french1979contemporary} and, though the relevance of the vast resulting literature to the present discussion is somewhat  indirect, it is felt to be worthy of mention nonetheless.  
The central point has been to question what is indicated by using a name, whether names are  essential and, if not, what they might be replaced with. Now it should be noted that typically the  discussion was of named objects vs. definite descriptions, and upon reflection cooking recipes  cannot contain either definite descriptions or uniquely identifying names. If they did they would  become chronicles of one particular instance of cooking, for example describing what took place  in John’s kitchen yesterday, rather than the blueprint for a potentially infinite number of such  chronicles that we expect them to be. The idea of a name can be employed in a looser sense to  provide a reference to an entity, rather than attaching an enduring label to it—‘the carrot’ as  opposed to ‘Louis Mahon’.\footnote{See \citeauthor{evans1982varieties} (\citeyear{evans1982varieties}) for a comprehensive discussion of different sorts of  names and references.}  For example it is this looser sense at play in the technique of Named  Entity Recognition (discussed in Section \ref{subsec:current-lit}). However, to avoid confusion remainder of this  paper equates the term ‘name’ with proper names, such as ‘Louis Mahon’.

With an acknowledgment of the difference between named entities and discourse referents in  mind, a parallel can be drawn between names vs. definite descriptions on one hand, and  discourse referents vs. states of affairs on the other. A particular connection might be seen with  Russell’s (\citeyear{russell1905denoting}) well-known analysis of ‘the present king of France’ as being anything that fits  the description of being the present king of France, or Searle’s claim that names are merely ‘pegs on which to hang descriptions’ (\citeyear{searle1958proper}, p172). These both suggest that a descriptions are more fundamental than named objects. Even Wittgenstein’s famously cryptic remark that  ‘the world is the totality of facts, not of things’ (\citeyear{wittgenstein1961tractatus}) can be broadly interpreted as stating that everyday objects, such as tables and chairs, supervene on states of affairs (see \cite{miller2017representation,copi2013essays,georgallides2015unclarity} for discussion of a more detailed interpretation). The key point of similarity  is that entities (whether named or not) are not essential, but rather to be used only when  convenient. In many circumstances it is indeed efficacious to be able to group multiple properties together. Consider for example the success of discourse segmentation methods based on changes in the discourse referents being used (\cite{fragkou2017applying, bayomi2015ontoseg}: if entities are in fact pegs holding together multiple  properties or ideas then it would be expected that identifying a change in the entities under  consideration would be an effective way of sectioning a text (more effective than, for example,  identifying a change in the verbs being used). However, in the cooking domain, whose central  purpose is to manipulate and recombine to produce a desired outcome, the same properties do not remain together for very long. Consequently entity based understanding, which often offers  a simplification, was felt to be less appropriate than a representation system in terms of states of  affairs.

%% file: content_production.tex
\section{Content Production} \label{sec:content-production}
In the terminology used in this paper, content production refers to the method of establishing all  the content that could potentially appear in a discourse. As described in section 1, FASTFOOD represents discourse content using descriptive strings and processes; thus the result of the work  described in this chapter is the creation of a database described in 
\begin{align*}
&can\_make &\gets &\text{ set of all descriptive strings recognised by FASTFOOD} \\
&skills &\gets &\text{ the set of all processes recognised by FASTFOOD} \,.
\end{align*}

This database can then be accessed by the algorithms described in the following Sections  whenever a user \text{input} is made.  
Section \ref{subsec:productive-method} gives an overview of what is required of the productive method. Sections \ref{subsec:foods-and-actions} and  \ref{subsec:synonyms-and-ghosts} introduce food classes, actions and synonyms, the three types of structure that form the essential components of the production of $can\_make$ and $skills$. 

\subsection{Overview of Productive Method} \label{subsec:productive-method}
\begin{figure}
    \centering
    \includegraphics[trim={0.1cm 0 0 0.1cm},clip, totalheight=0.8\textheight]{\detokenize{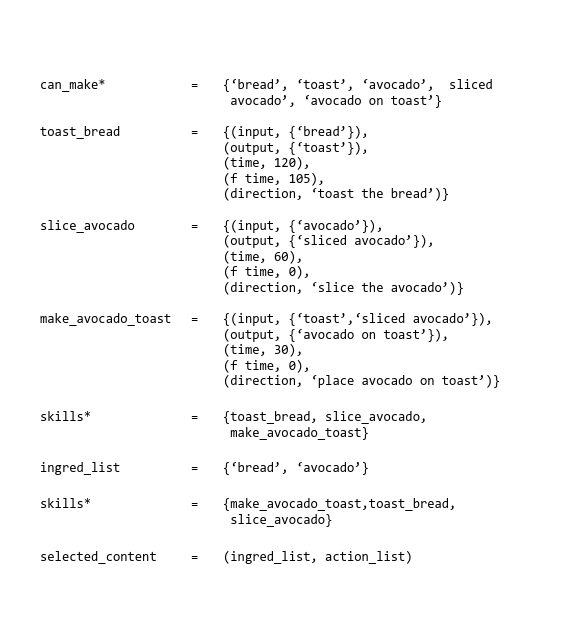}}
    \caption{Example extensional definition.}
    \label{wrong-def}
\end{figure}

One possible way of producing these sets would be to encode them extensionally. A simple hypothetical extensional definition, termed $skills$* and $can\_make$* is given in Figure \ref{wrong-def}, along with a demonstration of how it would be used to represent a recipe for ‘avocado on toast’. 

\begin{figure}
    \centering
    \includegraphics[trim={0.1cm 0 0 0.1cm},clip, totalheight=0.4\textheight]{\detokenize{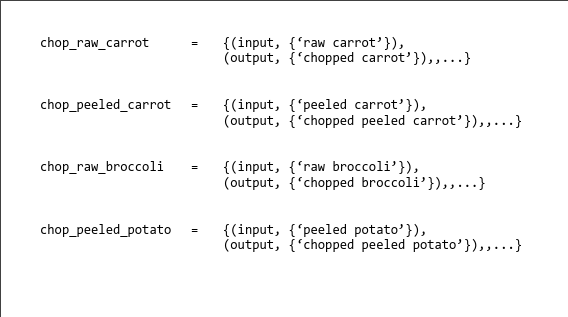}}
    \caption{Example of the many redundant processes that extensional definition requires.}
    \label{excessive-extensional}
\end{figure}

Extensional definition works in this example, but extending the domain to include multiple  recipes drastically increases the size of $can\_make$ and $skills$. Due to the specificity of each  descriptive string –‘chopped carrot’ describes a separate state of affairs to ‘chopped peeled  carrot’--and because processes are defined on specification of both an \text{input} and an \text{output}  string, a large number of similar processes would need to be written separately, as e.g. in Figure \ref{excessive-extensional}.  
In its current version FASTFOOD recognises over 400 processes ($can\_make$ has order greater  than 400) and any expansion would further increase this number. Therefore manual encoding is  not a feasible approach.  
Instead, the discourse content is produced intensionally by the interaction of several types of  structure, as described in the following two sections.

\subsection{Food Classes and Cooking Actions} \label{subsec:foods-and-actions}
This section discusses two types of structure used as part of this intensional production of  $can\_make$ and $skills$. These structures are cooking actions and food classes.  
Intuitively, cooking actions can be thought of as verbs and food classes as nouns. The current  version of FASTFOOD contains 8 cooking actions, including chop, boil, mash and soak; and a number\footnote{This number is different at different points in the production phase, see Figure 2.2.} of food classes including broccoli, coconut milk, potato and pasta. The  formation of $skills$ requires a specification of how these interact to obtain meaningful processes. The term ‘meaningful’ is used to describe a process that it is pertinent for FASTFOOD to consider; this excludes those that may be called physically impossible, such as chopping coconut milk along with those that may be possible but are highly impractical, such as soaking pasta.

A food class and cooking action are called compatible if they can together produce a meaningful  process. Syntactically, each action is a set of processes of a similar form. For example, part of the form for chop is that the \text{output} is ‘chopped’ concatenated with the \text{input}. These sets are  formed separately and then their elements are all added to $skills$. There are some differences  between actions, such as the inclusion of ‘oil’ in the \text{input} of frying and ‘boiling water’ in  the \text{input} of boiling, however the structure is essentially the same for each.  
The general form for chop is given in (2.1).
\begin{grammar} 
<chop-process> ::= <chop-\text{input}-pair> <chop-\text{output}-pair> <chop-time-pair> <chop-ftime-pair> <chop-direction-pair> 

<chop-\text{input}-pair> ::= ‘\text{input}’ <choppable-food>  

<chop-\text{output}-pair> ::= ‘\text{output}’ ‘chopped’ <choppable-food>  

<chop-time> ::= ‘time’ integer 

<chop-ftime> ::= ‘\text{f time}’ 0 

<chop-direction> ::= ‘direction’ ‘chop the’ <choppable-food> (2.1)
\end{grammar}

The element referred to as ‘choppable-food’ must, as per the syntax of a process in (1.1), be a  descriptive string. However, some descriptive strings would produce non-meaningful processes. Thus, it is specified which descriptive strings are compatible with which  actions.  
Rather than define a separate set of suitable descriptive strings for each cooking action, a single  set of descriptive strings is used, each accompanied by an indication as to which cooking actions it is compatible with. For example, the descriptive string ’avocado’ is compatible with chop and mash, but not with boil. This information is encoded by a tuple containing a number of  Booleans. Each Boolean indicates compatibility with each of the cooking actions; the tuple  containing ‘avocado’ would have elements indicating compatibility with chop and mash, and others indicating incompatibility with soak and boil. These tuples are termed food classes. 
Based on what has been said so far, food classes are a descriptive string and 8 Booleans (indicating True or False for each of the 8 cooking action). However, two modifications to this  characterisation are necessary. Firstly, a binary True/False as to whether a given descriptive  string is compatible with a given cooking action misses the fact that the same cooking action may take more or less time depending on the descriptive string. Having True for boil in both  ‘lentils’ and ‘spinach’ doesn’t allow for the fact that boiling spinach is considerably quicker  than boiling lentils. So, instead of True or False, the element is either False or an integer, the  latter specifying the ‘time’ value of the process that will be produced. This element is referred to as an indicator. 
Secondly, some sets of food classes are similar, both in their descriptive string and in their  indicators, and it is therefore efficient to define them off the back of one another. For example  the food classes for ‘peeled carrot’ can be defined from that for ‘raw carrot’ by replacing ‘raw’ with ‘peeled‘ in the descriptive string, and changing the peel indicator to False (to  encode the fact that you cannot peel a carrot once it has already been peeled). This is done by  splitting the descriptive string into two parts, called the root and the state. The full syntax for food classes is shown in (2.7).  
 
\begin{grammar}
<foodclass> ::= <root> <state> <Indicator> <Indicator>  <Indicator> <Indicator> <Indicator> <Indicator> <Indicator> <Indicator> 

<indicator> ::= False | integer (2.7)
\end{grammar}  

For each cooking action, a function is defined that acts on a food class, modifies the state and  certain indicators and produces a new food class. So the function for peel changes the state to  ‘peeled’ and changes the peel indicator to False. The formal description of these functions is  given in Algorithm \ref{alg:create-new-food-class}. The descriptive string of a food class formed by combining its root and  state will be termed its ‘description’. 
\begin{equation} \label{eq:description-def}
f[\text{description}] \gets \text{ set of all descriptive strings recognised by FASTFOOD} 
\end{equation}

Then, an initial set of food classes, foods, is specified extensionally (20 in the present version  of FASTFOOD) and Algorithm \ref{alg:produce-content}, which produces descriptive strings to  be placed in $can\_make$ and processes to be placed in $skills$, adds many more food classes to  foods along the way. 
First, the algorithm adds the descriptive strings for each of these food classes to $can\_make$. Then foods is iterated through and for each food class f: the set of cooking actions is iterated through,  and if f is compatible with a food class c, then a process is produced based on properties of both  f and c. The \text{output} of each new process forms the basis for the production of a new food class,  which is then added to foods. Thus, there is a recursive production where food classes combine  with cooking actions to produce processes and also new food classes.\footnote{In theory this recursion could run forever. However, the changes in indicators are designed in such a  way that this never happens. Specifically, every function sets the Boolean corresponding to itself as  False (Chop sets choppable = False, Boil sets boilable = False etc.) and once it is changed it is never  changed back. Thus the number of iterations for each food class is limited by the number of actions, in  this case 8. This is done by restraint, not by constraint; it is a factor which must be borne in mind when  defining the effects of the cooking actions.} Finally, the \text{output} of  every element of $skills$ is added to $can\_make$. 
At this point in the production phase, $skills$ includes all processes formed from the interaction  of the appropriate foods classes and cooking actions, and $can\_make$ contains all descriptive  strings that are \text{input}s or \text{output}s to these processes. These are then added to by the structures  discussed in Section \ref{subsec:synonyms-and-ghosts}.

\begin{algorithm}
\begin{algorithmic}
\caption{Formal description of functions used to create new food classes, f[\text{choppable}], f[\text{boilable}] and f[\text{fryable}]  refer respectively to f’s indicators for chopping, boiling and frying.} \label{alg:create-new-food-class}

\Function{chop}{$x$} 
\State $x \gets $ a food class 
\State $chopped\_x \gets x $ will be updated to show effects  of chopping
\State $chopped\_x[\text{state}]  \gets $ ‘chopped’
\State $chopped\_x[\text{choppable]} \gets False$ 
\State \Return chopped\_x 
\EndFunction

\Function{boil}{x} 
\State $x \gets $a food class
\State $boiled\_x \gets x $ will be updated to show effects  of boiling
\State $boiled\_x[\text{state]} \gets $ ‘boiled’
\State $boiled\_x[\text{boilable]} \gets False $
\State \Return boiled_x 
\EndFunction

\Function{fry}{x} 
\State $x \gets $a food class 
\State $fried\_x \gets x $ will be updated to show effects of frying
\State $fried\_x[\text{state]} \gets $‘fried’
\State $fried\_x[\text{fryable}] \gets False$ 
\State $fried\_x[\text{boilable}] \gets False$
\State \Return fried\_x 
\EndFunction

\end{algorithmic}
\end{algorithm}

\begin{algorithm}
\caption{Formal description of the algorithm used to produce many descriptive strings and processes, which are  then stored in can_make and skills, f[choppable[, f[boilable] and f[fryable] refer respectively to f’s  indicators for chopping, boiling and frying.} \label{alg:produce-content}

\begin{algorithmic}
\Function{produce_content}{foods}
\State $foods \gets$ predefined set of food classes
\State $chop \gets \emptyset$ , will be updated to contain all  chopping-related processes
\State $boil \gets \emptyset$, will be updated to contain all  boiling-related processes 
\State $fry \gets \emptyset$, will be updated to contain all  frying-related processes 
\State $skills \gets \emptyset$, will be updated to contain all chopping- boiling- and frying-related processes 
\State $can\_make \gets \emptyset$, will be updated to contain  recognised descriptive strings 
\For {each $f$ in foods }
\State add $f[\text{description}]$ to $can\_make $
\EndFor 
\For {each f in foods }
\If {$f[\text{choppable}$] is an integer }
\State $chopped\_f \gets chop(f) $
\State $chop\_f \gets (\text{input}, f[\text{description}]), (\text{output}, chopped\_f[\text{description}),$ \\ $(\text{time}, f[\text{choppable}), (\text{f \text{time}}, 0), (direction, \text{‘chop the `$[\text{root})$} $
\State add $chopped\_f$ to foods 
\State add $chop\_f$ to $chop$
\EndIf
\If {$f[\text{boilable}]$ is an integer }
\State $boiled\_f \gets boil(f) $
\State $boil\_f \gets (\text{input}, {f[\text{description}], \text{‘boiling water’}}), (\text{output}, boiled\_f[\text{description}]),$ \\ 
$(\text{time}, f[\text{boilable}]), (\text{f \text{time}}, f[\text{boilable}] - 30), (direction, \text{‘boil the $f[\text{root}]$ for $f[\text{boilable}]$`)}$
\State add $boiled\_f$ to foods 
\State add $boil\_f$ to $boil$
\EndIf
\If {$f[\text{fryable}]$ is an integer }
\State $fried\_f \gets fry(f) $
\State $boil\_f \gets (\text{input}, f[\text{description}]), (\text{output}, fried\_f[\text{description}]),$ \\
$(\text{time}, f[\text{fryable}]), (\text{f \text{time}}, f[\text{fryable}] - 120), (direction, \text{‘fry the $f[\text{root}]$ for $f[\text{fryable}]$`})$
\State add $fried\_f$ to foods
\State add $fry\_f$ to $fry$
\EndIf
\EndFor 
\State $skills \gets chop \cup boil \cup fry $
\EndFunction
\end{algorithmic}
\end{algorithm}

\subsection{Synonyms and Ghosts} \label{subsec:synonyms-and-ghosts} 
\begin{figure}
    \centering
    \includegraphics[trim={0.1cm 0 0 0.1cm},clip, totalheight=0.23\textheight]{\detokenize{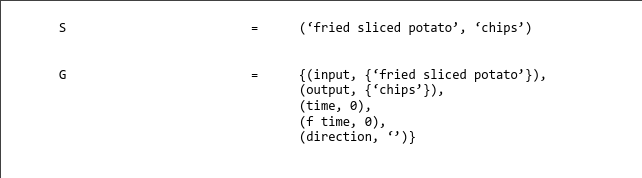}}
    \caption{Example of a synonym and the corresponding ghost process, for `chips' and `fried sliced potato'.}
    \label{fig:chips-syn-ghost}
\end{figure}

The ultimate goal of FASTFOOD is to receive a description of a dish as \text{input} and then \text{output} a recipe for that dish. In order to accomplish this, it must be able to understand the \text{input} description, i.e. the \text{input} description must an element of $can\_make$. The production method  described in Section \ref{subsec:foods-and-actions} produces only those descriptive strings constructed from combinations  of the given cooking actions and food classes, e.g. ‘chopped carrot’, ‘mashed chickpeas’;  and this excludes the likes of ‘hummus’, ‘ratatouille’ or ‘chips’. These latter sorts of description,  unlike ‘chopped carrot’ have nothing in their linguistic structure to indicate their physical  structure. The production methods of Section \ref{subsec:foods-and-actions} would apply the cooking action for frying to  the descriptive string ‘sliced potato’ to produce ‘fried sliced potato’; but a human  would instead use the description ‘chips’. Based on Section \ref{subsec:foods-and-actions} FASTFOOD would be unable to  understand this human description. 
The solution taken is to encode these anomalous descriptions, like ‘chips’, along with a  corresponding one that could be understood by FASTFOOD, ‘fried sliced potato’. Such a relation  between two strings, one of which is a descriptive string produced by the algorithm in figure  2.2, is defined syntactically in (2.8).
\begin{grammar}

<synonym> ::= <descriptive-string> <descriptive-string> (2.8)
\end{grammar}
For each synonym $(x,y)$ a process is produced which has $x$ as \text{input}, $y$ as \text{output}, zero time, zero \text{f time} and empty direction. These processes are termed ghost processes and are defined  syntactically in (2.9). 
\begin{grammar}
<ghost-process> ::= <\text{input}-pair><\text{output}-pair> <ghost direction> <ghost-time> <ghost-f-time> 

<\text{input}-pair> ::= ‘\text{input}’ <descriptive-string> 

<\text{output}-pair> ::= ‘\text{output}’ <descriptive-string> 

<ghost-direction> ::= ‘direction’ ‘’ 

<ghost-time> ::= ‘time’ 0  

<ghost-f-time> ::= ‘\text{f time}’ 0  (2.9)
\end{grammar}
In the case of ‘chips’, this results in Figure \ref{fig:chips-syn-ghost}. 
A ghost process with \text{input} x and \text{output} y, allows FASTFOOD to simplify the search for a recipe  for y to the search for a recipe for x. On a request for a recipe for chips FASTFOOD would, in  virtue of G, interpret ‘chips’ as being capable of being made from ‘fried sliced potatoes’,  and then begin searching for a recipe for the latter. \footnote{A more precise function of these ghosts can be understood in the context of the content selection  algorithm (Section \ref{sec:content-selection} and the worked example (Section \ref{sec:worked-example}).} G would initially be included in the  discourse content, but would be removed after the content selection stage (see Section \ref{sec:content-selection}) and  would make no appearance in the final generated \text{output}.  
Ghosts also provide a way of allowing for multiple methods of making the same dish, though  this is an underdeveloped aspect of the current version of FASTFOOD. Chips may also be made by baking sliced potatoes, and so the synonym S’ and the ghost G’ are also defined. 
\begin{figure}
    \centering
    \includegraphics[trim={0.1cm 0 0 0.1cm},clip, totalheight=0.29\textheight]{\detokenize{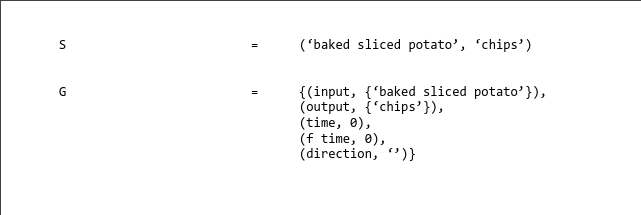}}
    \caption{Example of how synonyms and ghosts can allow for multiple ways of making the same dish: baking or frying potatoes to make chips.}
    \label{fig:chips-syn-ghost2}
\end{figure} 
If the content selection algorithm (Section \ref{sec:content-selection}) were to select $G$ then the final generated discourse  would be a recipe for chips by frying, whereas if it were to select $G’$ the recipe would be by baking. A possible extension of FASTFOOD would be to choose the most appropriate recipe by  requesting additional user \text{input}, i.e. asking the question ‘fried or baked?’. However, in the  current version the selection is made at random.  
The existence of both Figure \ref{fig:chips-syn-ghost} and Figure \ref{fig:chips-syn-ghost2} gives rise to the question of whether ‘chips’ and ‘fried  sliced potato’ are in fact synonymous after all. If synonymy is an equivalence relation, and both $S$ and $S’$ both describe such relations, then this would seem to imply the spurious conclusion  that ‘fried sliced potato’ is a synonym of ‘baked sliced potato’.\footnote{This concern is purely semantic. Synonyms are ordered pairs—ghost processes do not exist in both  directions—and so it would not be possible to use ‘chips’ as an intermediary and employ both S and S’ to conclude that one can make ‘fried sliced potato’ from ‘baked sliced potato’ or vice versa. 13 See \cite{cruse2002hyponymy} for a detailed analysis of the hyponym-hypernym relationship.}
From one perspective, the semantic relation between two elements of a synonym, is not one of  synonymy at all, but rather hyponymy-hypernymy. Two words that share the same or a very  similar meaning are said to be synonymous, e.g. ‘coriander’ and ‘cilantro’; whereas if the  meaning of $A$ is a subset of the meaning of $B$, e.g. ‘coriander’ and ‘herb’, then $A$ is said to be a  hyponym of $B$, and $B$ a hypernym of $A$. Viewed this way, synonyms and ghosts are a method  of expanding the vocabulary of FASTFOOD, e.g. to include ‘chips’, by specifying ‘chips’ as being a hypernym of an already understood description. Interestingly, there has been theoretical work expounding the importance of syntagmatic relationships such as hypernyms and hyponyms in the teaching of new vocabulary (Gao and Xu, 2013; Changhong, 2010); however discussion of  such work is outside the scope of the present paper. Treating ghosts as a method of vocabulary  teaching and weakening the corresponding syntagmatic relation from synonymy to hyponymy-hypernymy is one feasible interpretation which avoids the spurious conclusion that ‘fried sliced  potato’ is synonymous with ‘baked sliced potato’.

Another feasible interpretation is to view ghosts as instructions specifying which descriptive strings can be replaced with which others—as will be seen in Section 3.1, a ghost process is  selected for a recipe when the algorithm replaces a search for its \text{output} with a search for its \text{input}: if asked for a recipe for ‘chips’, G would allow FASTFOOD to replace ‘chips’ with  ‘fried sliced potato’, and instead search for a recipe for the latter. This amounts to adopting  a practical understanding of synonymy based on the ability to substitute one term for another,  and touches on some prior work based off such a definition. By employing various paraphrase  extraction techniques to a number of online corpora, \citeauthor{ganitkevitch2013ppdb} (\citeyear{ganitkevitch2013ppdb}) have constructed a large database of paraphrases. This database has allowed a  practical investigation of synonymous pairs based on the assumption that synonyms are words  that can act as paraphrases for each other (\cite{preotiuc2016discovering, pavlick2015inducing, koeva2015paraphrasing}. However, as noted by \citeauthor{koeva2015paraphrasing} (\citeyear{koeva2015paraphrasing}) synonymy understood in  terms of paraphrasing doesn’t always bear the formal properties of an equivalence relation—for  example ‘brain’ and ‘encephalon’ may be thought of as synonymous, and ‘brain’ could likely  be a substitute for ‘encephalon’ in the medical literature, but the reverse substitution could be  incongruous in less formal discourse, and so the relation does not bear the symmetrical property.

\begin{figure}
    \centering
    \includegraphics[trim={0.1cm 0 0 0.1cm},clip, totalheight=0.18\textheight]{\detokenize{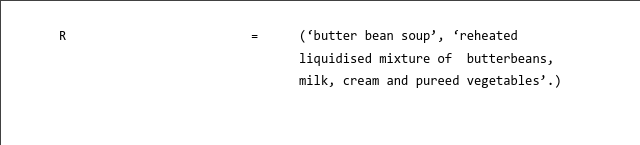}}
    \caption{Example of how specifying a synonym to describe the details of the dish you want to make.}
    \label{fig:recipe-synonym}
\end{figure}

By understanding synonymy in terms of paraphrasing, ‘chips’ could be interpreted as  synonymous with the two descriptive strings above, without facilitating the spurious conclusion  that these two strings are synonymous to each other.  
Certain synonyms are built in to FASTFOOD from the beginning—‘hummus’,  ‘chips’,’guacamole’,’pancakes’ etc., but often such a synonym may be required of the user, for  example if a recipe for ‘butter bean soup’ were requested, then the user would be required to add  a synonym like that in Figure \ref{fig:recipe-synonym}. 

It may seem a shortcoming of the system that a user would have to specify much of the essential  information, but there are two points in response to this. Firstly, as discussed, synonyms can be  interpreted as teaching FASTFOOD new vocabulary, so repeated use would improve its function:  a more experienced version of the program would already know several synonyms for (and hence, recipes for) ‘butter bean soup’\footnote{FASTFOOD currently chooses randomly between multiple recipes for the same dish,  but an extension of the system would allow a more intelligent choice, such as one guided by requesting  further user \text{input}.}. Secondly, the question ‘what exactly do you mean by  butter bean soup?’ is exactly the sort of question a menu reader, or personal chef, might ask. The description ‘butter bean soup’, like ‘chips’ is ambiguous in natural language, and the hesitation of FASTFOOD is an appropriate response to this ambiguity. 

The production method of Section \ref{subsec:foods-and-actions} produces a large number of descriptive strings of various  precise forms. However, the \text{input} made by a user may not be of such a form, and it is essential  that FASTFOOD understand the user \text{input}. Two types of structures, synonyms and ghost  processes, are defined to help ensure this understanding. These structures touch on two  theoretical issues; one relating to the role of syntagmatic relations in vocabulary learning, and  the other to differing understandings of the synonymy relation. After the production of the ghost  processes, they are added to $skills$, and $can\_make$ is updated accordingly.  

The result of the methods of this chapter are the creation of a database consisting of a set of  processes, called $skills$, and a set of descriptive strings, called $can\_make$. This database can  be stored and then accessed by the algorithms described in the following chapters whenever a  user \text{input} is made. These content production methods need only be run when FASTFOOD is  updated; i.e. on the addition of new food classes, cooking actions, or synonyms. 

%% file: content_selection.tex
\section{Content Selection} \label{sec:content-selection}
The algorithm in this section requires the following for operation:
\begin{itemize}
    \item the database produced by the methods of Section \ref{sec:content-production}, consisting of $can\_make$ and  $skills$ 
    \item a user description of the desired dish in a form understood by FASTFOOD, i.e. a  descriptive string that is an element of $can\_make$ 
    \item  user specified subset of $can\_make$, called $supplies$, corresponding to all foods  the user has at home 
\end{itemize}

The algorithm can then select which elements of $can\_make$ are required as ingredients, and  which elements of $skills$ are required to transform the ingredients into the desired dish. If the  ingredients are not all in $supplies$, then the program terminates with output ‘insufficient  ingredients, you need: $A1, A2,..., An$’ where $\{A1, A2,..., An\} = ingred\_list\setminus supplies$. The  formal syntax of a selected discourse content is given in (1.2). 
Section \ref{subsec:selection-algorithm} discusses the algorithm for selecting both the subset of $supplies$ that will form the ingredients list, and the subset of $skills$ that will collectively transform the ingredients list into  the desired dish. 
Section \ref{subsec:comparison-with-ATP-and-parsing} relates this algorithm to a procedure known as parsing and, in more detail, to well established work in the field of Automated Theorem Proving.  

\subsection{Selection Algorithm} \label{subsec:selection-algorithm} 
Equipped with $skills$ and $can\_make$ derived from the content production phase (Section \ref{sec:content-production}) and  $supplies$ as specified by the user, the Algorithm \ref{alg:select-content} can select the appropriate content for a recipe for the desired dish.

\begin{algorithm}
\caption{Algorithm that selects the appropriate ingredients (descriptive strings) and steps (processes) to form a  recipe for a dish requested by the user. } \label{alg:select-content}
\begin{algorithmic}

\Function{select\_content}{dish}
\State $ingred\_list \gets \emptyset$, will contain ingredients that are in $supplies$
\State $needed \gets \emptyset$, will contain ingredients that are not in $supplies$
\State $action\_list \gets \emptyset$, will contain all processes required to make the dish 
\State $looking\_for \gets $list containing only the dish, items will be removed from  this list by (a) finding them in $ingred\_list$ (b) placing them in needed or (c) finding a process for which they are the output and replacing them with the corresponding inputs 

\While {$looking\_for \neq \emptyset$ }
\For {each $item \in looking\_for$ }
\If {$item \in supplies$ }
\State remove $item$ from $looking\_for$ 
\State add $item$ to $ingred\_list$ 
\ElsIf {$item$ is not in $can\_make$} 
\State remove $item$ from $looking\_for$ 
\State add $item$ to $needed$ 
\Else
\For {each $p \in skills$ }
\If {item is an output of $p$ }
\State remove $item$ from $looking\_for$ 
\State add the inputs of $p$ to $looking\_for$ 
\State add $p$ to $action\_list$ 
\State break
\EndIf
\EndFor
\EndIf
\EndFor 
\If {$needed == \emptyset$ }
\State \Return $ingred\_list$, $action\_list$  
\Else
\State Print 'Insufficient ingredients, you need:' 
\For {each $x \in needed$ }
\State Print $x$ 
\EndFor 
\EndIf
\EndWhile
\EndFunction
\end{algorithmic}
\end{algorithm}

First, the descriptive string for the desired dish is set as the sole element of the list $looking\_for$.  The following is then performed until this list is empty: if the first element of the list is in  $supplies$ then it is removed and added to $ingred\_list$; otherwise, if the first element of the  list is not in $can\_make$ then it is removed and added to needed; otherwise, a process is found  with the first element of the list as output (a failure of the previous condition guarantees that such a process exists), the element is removed and the inputs of the process are added to the end of the list, the process is added to $action\_list$.  
If, after this loop, needed is non-empty then the dish cannot be made with the current supplies,  and so the program terminates and the elements of needed are printed, preceded by the directive  ‘Insufficient ingredients, you need:’. Otherwise $ingred\_list$ and $action\_list$ are returned to the content organisation phase (Section \ref{sec:content-organization}).  
Section \ref{subsec:synonyms-and-ghosts} exposited ghost processes and mentioned that, in addition to their primary function  of helping to ensure the user input is understood, they present a method for dealing with multiple ways of making the input dish. However, it can be seen in Algorithm \ref{alg:select-content} that the current content selection algorithm does not support this: only a single selected content is returned for each user  input. The desired extension could potentially be achieved as follows. For each element $X$ in  $looking\_for$ that is not in $can\_make$, search through through $skills$ and find every process that has $X$ as output (the current version only finds one such process), and then consider a separate case\footnote{Of course this is an incomplete specification. Further information would be required on how these  separate cases were considered.} for each such process. Each case would lead to a different selected content, which  would correspond to different recipes for the desired dish.  
In the absence of such an extension, the content selection phase acts on the produced database  of $can\_make$ and $skills$, along with the user input describing the desired dish and the foods on  hand, and returns a single selected content consisting of a set of descriptive strings called  $ingred\_list$, and a set of processes called $action\_list$.

\subsection{Comparison with Parsing and Automated Theorem Proving} \label{subsec:comparison-with-ATP-and-parsing}
This section conducts a comparison of the above method of content selection with two existing  types of algorithms: parsers and Automated Theorem Provers.  
Parsing is the general term for any technique that analyses a string of symbols in terms of its  role in a Formal Grammar. A Formal Grammar, as it is considered here, is a set of rewrite rules,  which describe how certain symbols can be rewritten as others, along with a privileged start  symbol from which the rewriting process may begin. A subset of these symbols are defined to  be terminal, and the set of all generated strings composed purely of terminal symbols is defined  as the language of the formal grammar.\footnote{For discussion of formal grammars see \cite{meduna2014formal, chomsky1956three}. } A prototypical instance of parsing is to determine  whether a given string is part of a given language, and if so to identify the production rules from which it can be generated. For example, treating natural language as a formal grammar, a  sentence can be parsed to determine whether it is grammatical and, if so, to produce a diagram,  sometimes called a parse tree, detailing its grammatical structure .
\footnote{Describing natural language is in fact the original use of Formal Grammars and can dated back to the  5th century BC, when the Indian linguist Panini used a Formal Grammar to describe grammatical  properties of Sanskrit \protect{\cite{bronkhorst2001panini,cardona1997panini}}. Over 2000 years later, Noam Chomsky (1956,  1957, 1963) proposed a rigorous system for understanding English syntax in terms of Formal Grammars;  and there have been many consequent developments in techniques for parsing natural language sentences,  \protect{\cite{collins2003head,tomita2013efficient}}.}
The similarity with the SELECT_CONTENT algorithm Algorithm \ref{alg:select-content} can be seen by equating  production rules to processes, terminal symbols to the descriptive strings in $supplies$, and non terminal symbols to all other descriptive strings. Both techniques aim to establish a connection  between a given element (dish/start symbol) and a given set of elements (subset of  $supplies$/string of terminal symbols) by applying certain transformations (processes/rewrite rules).  
 
However they are not identical for two reasons. Firstly, the set of symbols in a string is ordered,  whereas the descriptive strings in $ingred\_list$ are not—the parse of a natural language  sentence would change if the order of the words was changed, but the same is not true with  respect to the ingredients list of a recipe. This difference is worthy of note, but it turns out not  to be insurmountable. One could use a Formal Grammar that did not discriminate based on order:  a Formal Grammar such that for every permutation $P$ and rewrite rule\footnote{Many common Formal Grammars, for example Context Free Grammars, only include rewrite rules  containing a single non-terminal on the left hand side. However, there is no reason to restrict the present  discussion beyond the more general case. }
\[
a_1, \dots, a_n \rightarrow b_1, \dots, b_n\,,
\]
there  also exists a rule 
\[a_1, \dots ,a_n \rightarrow P(b_1, \dots, b_n)\,.
\]
Indeed, there has been significant work on applying  parsing techniques to unordered (or only partially ordered) domains such as image recognition \cite{hongxia2008efficient, elliott2013image, zhao2010rectilinear} and narrative comprehension of stories (\cite{zhang1994story} and the references therein).  
A more substantial barrier to analysing SELECT_CONTENT as a form of parsing is a fundamental  difference in their input and output. In a sense, the two proceed in opposite directions: parsing  determines, for a given string, whether it can be derived by applying various production rules to  a starting symbol, and if so it identifies these production rules; FASTFOOD determines, for a  given dish, whether it can be made by application of various processes to a collection of  descriptive strings, and if so it identifies these processes. The former receives a string of  terminals as input and attempts to relate this string to a single pre-defined (start) symbol, the  latter receives a single symbol as input and attempts to relate it to a pre-defined set of non terminals. This difference should not be confused with that between bottom up and top-down  parsing. Top-down parsing, as in \cite{earley1970efficient} or \cite{frost2007modular} applies  rewrite rules to the start symbol to attempt to derive a given string; bottom-up parsing, as in  \cite{kasami1966efficient} or \cite{hall2003language} applies rewrite rules (or, more precisely, their  inverses) to a given string to attempt to derive the start symbol;\footnote{See \cite{aho1986principles} for an overview of different parsing methods and \cite{rau1988integrating} for an attempt to integrate the two approaches.} but in both cases, the string is the input, whereas the corresponding structure of SELECT_CONTENT —the ingredients list—is in  the output.  
As a related point, the start symbol in Formal Grammars is unique, but the corresponding  element in SELECT_CONTENT is the input query, the dish for which FASTFOOD is being asked to  compute a recipe, and any recipe generation program should be able to work with more than one  dish. In light of these differences, the task performed by SELECT_CONTENT can be understood  roughly, but not be exactly, as a form of parsing.

Another fruitful comparison is that with Automated Theorem Provers (ATP). An ATP is an  algorithm that seeks a method of proving a formula$ P $from a known list of axioms and  implicational rules. In the case of a backwards chaining ATP, the basic method for proving$ P $is  to search for an implication R with consequent P, thus reducing the search for$ P $to a search for  the antecedents of R, and continue to apply this method to these antecedents until the formulae  to be proved are all in the list of axioms. If this can be done then$ P $can be established as a  theorem provable by the axioms reached and the implicational rules used to reach them.\footnote{For a full exposition of the various approaches to Automated Theorem Proving, see \cite{geffner2013computational}} When  compared with the SELECT_CONTENT algorithm, the set of axioms corresponds to $supplies$, the  implications to $skills$, and$ P $to the desired dish. Deriving a proof of the formula$ P $corresponds  to generating a recipe for the dish. 
One important difference between ATP and SELECT_CONTENT is that the latter contains explicit  implications, ‘raw carrot’ $\rightarrow$ ’chopped carrot’, and the former implicational rules e.g.$(P\rightarrow Q)$. Implicational rules are more widely applicable than explicit implications. Indeed,  processes can only be applied if the consequent (output) exactly matches a specific proposition  to be proved, whereas implicational rules can describe an arbitrary number of implications. The  consequent of modus ponens for example, can be matched to any formula and so the search  could proceed indefinitely: proving $P_1$ could be reduced to proving $P_2$ and $(P_2\rightarrow P_1)$, then  proving $P_2$ could be reduced to proving $P_3$ and $(P_3\rightarrow P_2)$.\footnote{Such a possibility recalls Lewis Carrol’s famous discussion of the validity of Modus Ponens in a  conversation between Achilles and the Tortoise (1895).} Without additional constraints, there is no end to the number  of implications available to a theorem prover containing modus ponens, no point at which it will have exhausted all options. In the case of SELECT_CONTENT however, if a formula (descriptive string) doesn’t exactly match the consequent of any rule (isn’t in $can\_make$) then there are no  options left and the formula can be declared unprovable (the dish unmakeable).  
The difference is an advantage of SELECT_CONTENT. Formally, it means that given a set of foods and cooking actions, the set of makeable dishes is recursive, whereas it has long been known that the set of provable formulae is merely recursively enumerable \cite{godel1929vollstandigkeit}. Practically, it  means that SELECT_CONTENT is largely immune to the significant danger faced by ATP of being  led down blind alleys and failing to halt. Heuristics are necessary to guide ATP search in an  ‘intelligent’ direction, e.g. not continue to apply modus ponens as above. Indeed much of the  initial work of computer scientists in 1950s and 1960s concerned ATP and, working largely  within rigid systems such as Principia Mathematica, they enjoyed some success; but it was soon  realised that development of more effective heuristics would be necessary to move the enterprise  forward (see \cite{loveland2016automated} or \cite{bledsoe1984automated} for a discussion of the early developments in automated theorem proving). Such a situation has persisted: robust automated theorem proving  is still very much an open problem \cite{avigad2014formally}, and the bulk of the work in the  area is still devoted to intelligent proof guidance \cite{sutcliffe2001evaluating, urban2010evaluation}, with a growing trend to derive such guidance from machine learning (\cite{loos2017deep, schulz2015proof, bridge2014machine}.  In contrast, SELECT_CONTENT can show success in the absence of any search heuristics  whatsoever.

Another problem faced by Automated Theorem Provers is that the resulting proofs, when they  are obtained, are often extremely long and difficult for human readers to understand \cite{fontaine2011compression}. As a result, there have been  considerable efforts to increase the readability of such proofs, for example by removing  superfluous or repeated inferences \cite{fontaine2011compression}) or by graphically representing parts of  the reasoning process \cite{flower2004generating, stapleton2007automated}. The inferences made by mathematicians are  of a much higher order than the fundamental implicational rules such as modus ponens, a single  step in the head of a mathematician might correspond to thousands or even tens of thousands of  logical inferences\cite{fontaine2011compression}, and so a proof written in logical notation may be  exceptionally long. SELECT_CONTENT on the other hand, does not face such problems. A cooking  recipe is rarely more than a couple of pages: the average recipe has less than ten ingredients \cite{kinouchi2008non} and a brief search through 3 commercial recipe books \cite{flynn2016world,culleton1999simply,gavin1987italian} encounters nothing longer than 50 steps, with the vast majority being under 20.\footnote{ The longest encountered was a recipe for Mushroom and Leek Pie with Roast Potatoes and Cabbage \cite{culleton1999simply} p93.), which was analysed as containing 32 steps. Different analyses may  produce slightly different figures, but it is unlikely that any of these figures would be over 50; and it is  almost inconceivable that they would, as in the case of many ATP proofs, be in the thousands.} SELECT_ CONTENT is therefore unlikely struggle with overly long output.  

One potential issue with the treatment of recipe generation as theorem proving concerns resource  sensitivity: applying a process results in the deletion of the input, whereas, in classical logic,  applying a rule of inference does not delete the antecedent. So if the idea of chopping carrot  were to be expressed as ‘raw carrot’ implies ‘chopped carrot’, then we could use the  inference $A \land (A \implies B))\implies (A \land B)$ to incorrectly conclude that if there is raw carrot and we  can chop carrots, then we could produce a state of affairs in which there is both raw carrot and  chopped carrot. However this problem could be circumvented by a switch to a resource sensitive logic. Linear logic, for example, includes the notion of linear implication to describe cases in which the antecedent must be deleted in order to produce the consequent. There is a substantial  literature on the potential of ATP in linear logic \cite{galmiche2000connection, galmiche1995semantic,galmiche1994foundations}.  

The action of SELECT_CONTENT demonstrates that a simple backwards chaining ATP algorithm  can effectively select discourse content for a recipe from a predefined database. Moreover, this  method can be seen as particularly effective as two significant obstacles to ATP in  mathematics—unmanageable search spaces and excessively long proofs—are unlikely to present themselves here. 

A final point is that the two objects of comparison in this section, parsing and ATP, can  themselves be constructively compared. This comparison is not explored in any depth here, but  the characterisation of parsing as deduction has been conducted in detail by numerous authors \cite{shieber1995principles} since its formulation by \citeauthor{pereira1983parsing} in \citeyear{pereira1983parsing}. The same idea is the basis of the significant work employing logic programming to  parse natural language, \cite{mooney1996inductive, matsumoto1983bup, stabler1983deterministic}, logic programming being essentially a systematisation of simple ATP  methods \cite{lloyd2012foundations}. Both ATP and parsing, and particularly the former, have been seen to be similar to FASTFOOD’s  content selection algorithm; and readers wishing to, so to speak, complete the triangle, can  consult the comprehensive literature relating ATP and parsing to each other.

%% file: content_organization.tex
\section{Content Organization} \label{sec:content-organization}

\subsection{Overview of the Optimisation Problem} \label{subsec:optimization-problem-overview}
The algorithm in the previous section is capable of selecting the discourse content for a recipe  from a database (Section \ref{sec:content-production} given the necessary user input (Section \ref{sec:content-selection}). This selected content is  represented as two sets:
\begin{itemize}
    \item $ingred\_list$: a set of descriptive strings corresponding to the ingredients list  
    \item $action\_list$: a set of processes each transforming one descriptive string into  another, such that successive 
\end{itemize}
application of the processes in $action\_list$ to $ingred\_list$ will produce the desired dish 
The next phase is to organise this content, and the principles used to guide this organisation are  that the final recipe (a) be coherent, i.e. not call for an ordering of the steps that is physically impossible  such as straining the potatoes before boiling the potatoes (b) be as time efficient as possible, i.e. order and combine the steps in such a way that  the overall cooking time is as short as possible.
Fulfilling (a) requires a sensitivity to the fact that not all orders form coherent recipes. Some  processes can only be performed after the completion of others, e.g. one can only strain the  lentils after boiling the lentils. Such constraints occur when one process takes as input a  descriptive string that is in the output of another. The notation $requires(P_1,P_2)$ is used to  represent this relation. If neither process requires the other they are termed independent. An  order is called permissible if, for any two processes $P_1$ and $P_2$, $P_1$ requires $P_2$ only if $P_2$ precedes  $P_1$. 
\begin{definition} \label{def:requires}
For any two processes $P_1$, $P_2$, $requires(P_1,P_2)$ iff the input of $P_1$ and the output of $P_2$ have non empty intersection. 
\end{definition}
\begin{definition} \label{def:independent}
Two processes $P_1,P2$ are independent iff neither  $requires(P_1,P_2)$ nor $requires(P_2,P_1)$.
\end{definition}
\begin{definition} \label{def:permissible}
A list of processes $L$ is permissible iff for any  $P_1$, $P_2$ in $L$, $requires(P_1,P_2)$ implies $P_2$ precedes $P_1$ in $L$.  
\end{definition}
Thus part of the requirement in (a) is that the list into which the processes in $action\_list$ are  organised is permissible. 
The point in (b) is referred to as temporal optimisation, and there are two reasons such  optimisation is desirable. Firstly it demonstrates that the representation system is robust enough  to facilitate an inference of the sort that might be made by a human chef; and secondly, scope  for the application of optimisation procedures has been identified as a desirable feature of  representations used by smart kitchen appliances \cite{hamada2005cooking, reichel2011mampf}. The basis for temporal  optimisation is that some processes have times during their execution in which the cook is passive or free (the assumption is made that there is only one cook), for example while the  casserole is in the oven; and during these times one may squeeze in other processes which, if  performed in isolation, would add to the overall cooking time.  
The component of the organised content that relates to the processes, as opposed to the  ingredients list, is referred to as the recipe plan. Forming a recipe plan that will call for $P_1, \dots, P_n$ to be performed while performing P will be referred to as inserting $P_1, \dots, P_n$ into $P$ or  combining $P, P_1, \dots, P_n$, and the result will be referred to as a combination. The result of  ordering and combining the processes in $action\_list$ is referred to as a recipe plan. This  terminology is formalised in (4.2). 

\begin{grammar}
<organised-content> ::= <ingredients-list> | <recipe-plan> 

<ingredients-list> ::= <descriptive-string> |  <descriptive-string> <ingredients-list> 

<recipe-plan> ::= <process>| <combination> | <process> <recipe plan> |  <combination> <recipe-plan>

<combination> ::= ‘while’ <process> ‘,’ <process-list> 

<process-list> ::= <process> | <process> ‘and’  <process_list>  (4.2)
\end{grammar}

Just as there are restrictions on which orders are coherent, so too are there restrictions on which processes can be coherently combined. One condition for inserting $P_1$ into $P_2$ is that the cook  be free for long enough during $P_2$ for $P_1$ to be performed. A process is encoded as a $5$ element  associative array and two key-value pairs of this array have form (time, $M$) and (f time, $N$)  where $M$ and $N$ are non-negative integers. These key-value pairs respectively express the total time the process will take, and the amount of time during the process at which the cook is free.  
So by definition, combinations are only possible when the total time of the inserted processes is  less than or equal to the free time of the process into which they are to be inserted. If making  toast leaves the cook free for only $60s$, and chopping tomatoes takes $90s$, then it is incoherent to  combine making toast and chopping tomatoes.  
A second restriction is imposed by the requires relation. $requires(P_1,P_2)$ means $P_1$ must occur  after the completion of $P_2$, so they cannot occur at the same time. One cannot chop the broccoli  while boiling the chopped broccoli. The two conditions for the insertion of $P_1$ into $P_2$ are shown  in \eqref{cond:insertion}. 

\begin{condition} \label{cond:insertion}
$P_1[\text{time}] \leq P_2[\text{f time}]$, and $P_1$ and $P_2$ are independent
\end{condition}

Therefore another requirement of (a) is that no combinations are performed that violate these  two conditions. 
Temporal optimisation is performed by considering multiple organised discourses, calculating  the total time of each organised discourse by adding together the times of each process that  hasn’t been inserted into another, and then selecting the organised discourse with the least overall  time. It is contended that the precise way this is done is guaranteed to achieve an optimal  solution, i.e. there will exist no faster coherent organised discourse with shorter overall time;  however a proof of this contention is outside the scope of the present work.  
The content organisation algorithm contains two main components. The first generates all  permissible lists of the processes in $action\_list$, the second acts on each list to combine certain  processes. These two components are the respective subjects of the following two subsections. 
 
\subsection{FIND_ALL_LISTS} \label{subsec:find-all-lists}
Before the application of the algorithm to generate all permissible lists of the processes in  $action\_list$, any ghost processes (see Section \ref{subsec:synonyms-and-ghosts}) are removed. After doing this, the requires  relations in $action\_list$ must be updated to ensure no impermissible lists are generated. If $A, B, G$ are $3$ processes and $G$ is a ghost; and if $requires(A,G)$ and $requires(G,B)$ then when $G$ is removed, the relation $requires(A,B)$ must be added. This update procedure is decribed in Algorithm \ref{alg:remove-and-stitch}.  

\begin{algorithm}
\caption{Algorithm for removing ghost processes from $action\_list$ before application of FIND_ALL_LISTS.} \label{alg:remove-and-stitch}
\begin{algorithmic}
\Function{remove\_and\_stitch}{L}
\State $L \gets$ set of processes, some of which may be ghosts
\For {each item in $L$  }
\If {item[\text{time}] == 0 }
\State $new\_requirements \gets \emptyset$ that, will be updated to equal the set of all elements that item requires
\For {each item’ in $L$ }
\If {requires(item,item’) }
\State add item’ to new\_requirements 
\EndIf
\EndFor 
\For {each item’’ in $L$ }
\If {requires(item’’,item) }
\For {each req in new\_requirements }
\State $requires(item,req) \gets True $
\EndFor 
\EndIf
\EndFor 
\State remove item from $L$ 
\EndIf
\EndFor 
\EndFunction
\end{algorithmic}
\end{algorithm}

The input set of processes, in this case $action\_list$, is iterated through. For each ghost $G$ found,  the set of all processes that $G$ requires are placed in new_requirements; then all processes that require $G$ are updated to require everything in new_requirements; then $G$ is removed.\footnote{The necessity of such a specification can be seen as follows. Before the removal of $G$, permissibility  would require that $B$ precede $G$ and $G$ precede $A$, and thus $B$ would have to precede $A$. If $G$ were to be  removed and the requires relations not updated, there would be nothing to prevent $A$ preceding B, and this would result in an incoherent recipe. As discussed in Section \ref{subsec:synonyms-and-ghosts} semantically requires the input of $G$, and the fact that $requires(G,B)$ holds means the input of $G$ is in the output of $B$. Therefore $A$ semantically requires something in the output of $B$ and so any order in which $A$ precedes $B$ must be marked as  impermissible. This is ensured precisely by specifying that $requires(A,B)$.}
After this removal of ghosts, the algorithm for finding all permissible lists, called  FIND_ALL_LISTS, is run. A necessary precursor to this algorithm is Definition \ref{def:no-requirements}.  
\begin{definition} \label{def:no-requirements}
$no\_requirements(x,S)$ iff $X$ is a process and $S$ is a set of processes such that requires(x,s) does not  
hold for any element $s$ of $S$ 
\end{definition}
If $x$ is an element of the given set $S$ then $no\_requirements(x,S)$ is equivalent to saying  that $x$ can be the first element in a permissible order of $S$. Algorithm \ref{alg:all-possible-lists} shows how, using  an extension of this idea, FIND_ALL_LISTS can compute all permissible orders of a given  set of processes. 
It begins by adding to paths all elements that have no requirements in $S$ and so could come first  in a permissible order. Then it iterates through paths, replacing each element $x$, with all lists  $[x,y]$ where $y$ is a process that doesn’t require any other except for possibly $x$; then it again  iterates through paths, replacing each list $[x,y]$ with all lists $[x,y,z]$, where $z$ is a process that  doesn’t require any other except for possibly $x$ or $y$. It iterates this operation a number of times  equal to the cardinality of the input set. At this point paths contains all permissible orders as  desired.\footnote{It is possible to represent the steps in a recipe using a Directed Acyclic Graph. Figure \ref{fig:DAG-wo-strings} below is an  example of such a representation, and techniques for extracting DAG representations from natural  language recipes are discussed in Section \ref{subsec:current-lit}. A DAG, in turn, is equivalent to a partial  order, see \cite{kozen1992design} for a proof of this equivalence and a general discussion of partial orders. Thus  the above search for all permissible orders is equivalent to the search for all total orders consistent with a  partial order, and so FIND_ALL_LISTS is similar to the Kahn algorithm for topological sorting (\citeyear{kahn1962topological}). The  latter determines a single total order consistent with a given partial order, the algorithm here determines all total  orders consistent with a given partial order.}

\begin{algorithm}
\caption{Algorithm that determines all permissible lists of a given set of processes. \#S refers to the  cardinality of the set S} \label{alg:all-possible-lists}
\begin{algorithmic}
\Function{possible\_next}{S,L}
\State $S \gets $a set of processes
\State $L \gets $a list of processes in S
\State $possibles \gets  \emptyset$ will contain all permissible lists obtained by adding a further element of $S$ to $L$
\State remove all elements of $L$ from $S$ 
\For {each element $X$ of S: }
\If {no\_requirements(x,S)  }
\State add $X$ to possibles 
\EndIf
\EndFor 
\State replace each element y of possibles with the list $L$ + [y]  add each element of $L$ to S
\State \Return possibles
\EndFunction
\Function{find\_all\_lists}{S}
\State $paths \gets \emptyset$ that will contain all permissible lists 
\For { each element $x$ in S}
\If {no\_requirements(x,S) then }
\State add the list containing only $x$ to paths 
\EndIf
\EndFor 
\For {i in range(\#S-1) }
\For {each element z in paths: }
\For {each element $p$ inwith possible\_next(S,z)}
\State add $p$ to paths
\EndFor 
\State remove z from paths			
\EndFor 
\EndFor 
\State \Return paths 
\EndFunction
\end{algorithmic}
\end{algorithm}

\subsection{CONCURRENT_COMPRESSION} \label{subsec:concurrent-compression}
The second component of the content selection algorithm is referred to as CONCURRENT_ COMPRESSION. It receives as input a list $L$ of processes (note, $L$ is already coherent) and combines certain processes in this list. We say that process $p$ is ‘inserted into’ process $p'$ if the algorithm determines that $p$ should be performed during the passive time of $p'$. We adopt the convention that insertions are always performed from right to left. This does not miss any possibilities because all permissible orders are considered (as described in Section 4.2). The algorithm proceeds as follows. For each process $p$ in $L$, it forms another list concurrents containing all the processes directly following $p$ which may be inserted into $p$. It then checks  all elements $p'$ preceding $p$ in $L$ to see if $p$ may be inserted into $p'$. If so, it assigns future_insertee to such a $p'$ and removes all elements from concurrents that may interfere with this insertion—all elements that require future_insertee.\footnote{The purpose of this look ahead procedure is to ascertain whether a given combination will prevent another, more substantial one. The potential future insertion would definitely be more substantial, given the input ordering, because the process being inserted into has to have time strictly larger than that of the inserted one.} The formal description of the  total CONCURRENT_COMPRESSION algorithm is given Algorithm \ref{alg:concurrent-compression}. 
In Section \ref{subsec:optimization-problem-overview} an organised discourse was said to be coherent iff
\begin{enumerate}[i]
    \item it does not contain two processes $P_1$ and $P_2$ such that $P_1$ precedes $P_2$ in the list and  $requires(P_1,P_2)$ 
    \item it does not contain two processes $P_1$ and $P_2$ such that $P_1$ and $P_2$ are combined and  $requires(P_1,P_2)$ 
    \item it does not contain a process $P$ and a set of processes $S$ such that every element of $S$ has been inserted into $P$, and the sum of the total time of every element of $S$ is greater than the free time of $P$
\end{enumerate}

The output of CONCURRENT_COMPRESSION is guaranteed to be coherent in this sense.  
It accomplishes (i) by inserting neighbouring processes one at a time and ending the insertions  when an incompatible process is encountered. Thus only contiguous sublists of processes are combined  and so the order is not changed, i.e. there are no two processes $P_1$ and $P_2$ such that $P_1$ preceded  $P_2$ in the input and $P_2$ precedes $P_1$ in the output. The input list satisfies (i), and  therefore so does the output.  
 
\begin{algorithm}
\caption{Algorithm that iterates through a given list of processes and performs combinations under certain criteria.} \label{alg:concurrent-compression}
\begin{algorithmic}
\Function{concurrent\_compression}{L}
\State $L \gets $ list of processes
\For {each process $p$ in $L$, beginning from the end of the list}
\State $concurrents \gets $ empty list that will contain inserted processes 
\For {each process $p'$ to the right of $p$, moving right}
\If {$p'[\text{time}] \leq p[\text{f time}]$ and $p$ and $p'$ are independent }
\State add $p'$ to $concurrents$ 
\State $p[\text{f time}] \gets p[\text{f time}] - p'[\text{time}] $
\EndIf
\EndFor 
\State combine$(concurrents,P,L)$
\EndFor 
\EndFunction

\Function{combine}{concurrents,p,L}
\State $concurrents \gets $ a list of processes
\State $p \gets $ a process into which the processes in  concurrents will be inserted
\State $L \gets $ concatenation of $p$ and $concurrents$
\State $future\_insertee \gets  \emptyset$
\State $required\_f\_time \gets p$[\text{time}]
\For {each $p'$ preceding $p$ in $L$, moving left }
\If {$requires(p,p')$ or $requires(p',p)$ }
\State break 
\ElsIf {$p'[\text{f time}] \geq required\_f\_time $}
\State $future\_insertee \gets p'$
\State break 
\Else
\State increase $required\_f\_time$ by $p'$[\text{time}]
\EndIf
\EndFor 
\If {$future\_insertee \neq \emptyset$}
\For {each $q$ in concurrents }
\If {$q$ and $future\_insertee$ are not independent }
\State remove $q$ from $concurrents$ 
\EndIf
\EndFor 
\EndIf
\For {each $item \in concurrents$}
\State add the input of $item$ to the input of $p$ 
\State add the output of $item$ to the output of $p$ 
\State $x $ stands for all known processes
\If {$requires(item,x)$}
\State $requires(p,x) \gets True$
\EndIf
\If {$requires (x,item)$}
\State $requires(p,item) \gets True$
\EndIf
\State remove $item$ from $L$ 
\EndFor 
\State $concurrent\_instruction \gets $ concatenation of the ‘direction’ entry  of all processes in concurrents, with  ‘and’ occurring between each 
\State $p[direction] \gets$ ‘while’ + $p[\text{direction}]$ +  $concurrent\_instruction$
\EndFunction
\end{algorithmic}
\end{algorithm}

It accomplishes (ii) by making it an explicit condition for the combination of two processes that  neither requires the other.  
It accomplishes (iii) by making it an explicit condition for the insertion of $P_1$ into $P_2$ that  $P_1[\text{time}] \leq P_2[\text{f time}]$; and then inserting one at a time and, whenever $P_1$ is inserted into $P_2$,  reducing the free time of $P_2$ by an amount equal to the total time of $P_1$.  
Thus, for each permissible list of the processes in $action\_list$, CONCURRENT_COMPRESSION returns a coherent recipe plan. FASTFOOD then determines the total time of each of these recipe  plans by summing the times of all processes that have not been inserted into another, and selects  that with shortest total time.
It is contended that the temporal optimisation method is guaranteed to obtain an optimal solution,  that there will not exist a coherent recipe plan for the processes in $action\_list$ which has total  time less than that selected by FASTFOOD. However, a proof of this contention is outside the  scope of the present paper. It may be left to the reader to investigate based on the given  description of the algorithms. 

%% file: content_realization.tex
\section{Content Realization} \label{sec:content-realization}
As a result of the stages describe in Section \ref{sec:content-production}, \ref{sec:content-selection} and \ref{sec:content-organization}, the organised content is represented  by the two following structure:. 
\begin{itemize}
    \item $ingred\_list$: a set of descriptive strings corresponding to the ingredients of the recipe;
    \item $recipe\_plan$: an ordered list of processes some of which have possibly been combined, such  that when successively applied to the $ingred\_list$, the desired dish will be produced.

\end{itemize}
This content is then realised by printing  
\begin{itemize}
    \item the name of the dish 
    \item the total cooking time 
    \item the ingredients list 
    \item the instructions 
    \item the times during the recipe at which the cook is passive
\end{itemize}
The syntax of the realised discourse in FASTFOOD is given in Backus-Naur Form in (5.1).  
\begin{grammar}
<realised-discourse> ::= <recipe-title> ‘Time:’ <time> ‘Ingredients’ <ingred-list>  ‘Instructions’ < instruction-list> ‘Passive times:’ <passive-times> 

<recipe-title> ::= <descriptive-string> 

<ingred-list> ::= <ingred-list> |  <descriptive-string> <ingred-list> 

<instruction-list> ::= <instruction> |  <instruction>  <instruction-list> 

<passive-times> ::= <passive> |  <passive> <passive-times> (5.1) 
\end{grammar}

\begin{algorithm}
\caption{Algorithm for extracting temporal information from $recipe\_plan$. } \label{alg:extract-temporal-info}
\begin{algorithmic}
\Function{get\_passives}{l}
\State $l \gets $ a list of processes
\State $clock \gets 0$
\State $passives \gets \emptyset$, will contain all intervals of passive  time in the recipe
\For {each $x$ in $l$}
\State $clock \gets clock + x[time]$
\If {$x[\text{f time}] > 0$}
\State add (hms$(clock - x[\text{f time}])$, hms$(clock)$), $x[text{direction}]$) to  passives 
\EndIf
\EndFor 
\State \Return $passives, clock$ 
\EndFunction
\end{algorithmic}
\end{algorithm}

\begin{grammar}
<instruction> ::= <time> string 

<passive> ::= ‘from' <time> ‘to’ <time> ‘while’ string 

<time> ::= integer ‘hrs’ | integer ‘min’ | integer ‘secs' |  integer ‘hrs’ integer ‘min’ |  integer <min> integer ‘secs’ | integer ‘hrs’ integer ‘min’ integer ‘secs’
\end{grammar}

The times on each process are written in seconds, so this sum is converted to hours, minutes and  seconds before being printed. The total cooking time and the passive times can be computed by Algorithm \ref{alg:extract-temporal-info}. The function HMS is the converter from seconds to hours, minutes and  seconds.  
This iterates through each $p$ in $L$, using the variable $clock$ to keep track of the time at which $p$ ends.\footnote{Marking the end rather than the beginning amounts to the assumption that the passive portion of each  process will occur at the end of that process.} If $p$[\text{f time}]$ > 0$ then the triple $(clock – p[\text{f time}], clock, p[text{direction}])$  is added to passives. This information is realised by printing ‘Passive’, and then ‘from $x$ to $y$ while $z$’ for each element $(x,y,z)$ in passives.  
The recipe content is realised by printing the ingredients list, instructions, total time and passive  times. An obvious shortcoming of the simple realisation method can be found in the instructions  of the combined processes, those of the form ‘While $A$, $B$’. Here $B$ should be, but is not,  modified to the progressive aspect. For example, Figure \ref{fig:veg-dahl-recipe2} has ‘while boil the lentils, chop  broccoli’ instead of ‘while boiling the lentils, chop broccoli’. It is felt that a solution to this  shortcoming would not be overly difficult; nevertheless it is outside the scope of the present  work.  

%% file: worked_example.tex
\section{Worked Example: Generating Recipe for Vegetable Dahl} \label{sec:worked-example}
\begin{figure}
    \centering
    \includegraphics[trim={0.1cm 0.5cm 0 0.1cm},clip, totalheight=0.28\textheight]{\detokenize{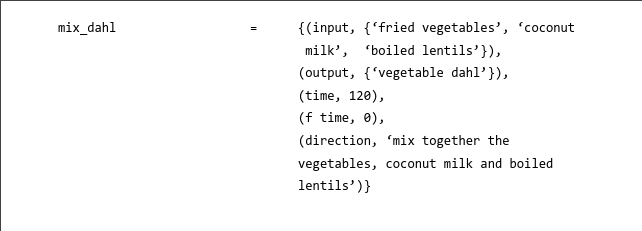}}
    \caption{Definition of the process required as the final step in cooking vegetable dahl.}
    \label{fig:mix-dahl}
\end{figure}

What follows is a worked example of how FASTFOOD would generate the recipe for the dish  described by the string ‘vegetable dahl’. The application of the algorithms discussed in Sections \ref{sec:content-production}, \ref{sec:content-selection}, \ref{sec:content-organization}, \ref{sec:content-realization}, and how they combine to produce a time efficient recipe is demonstrated. 
Before ’vegetable dahl’ can be used to generate a recipe, one must also input a specification  of what it is composed of in terms of descriptive strings already understood by FASTFOOD. The  specification amounts to defining what will be the final process in the recipe, the process that  produces the string ‘vegetable dahl’ itself, and in this case is given by the process in Figure \ref{fig:mix-dahl}. 

The user would also specify what foods are at hand, by defining the descriptive strings in  $supplies$ (see Section \ref{subsec:selection-algorithm}. For the purposes of this example it will be assumed that $supplies$ contains the descriptive strings ‘coconut milk’, ‘raw carrot’, ‘raw broccoli’,  ‘lentils’ and ‘water’.  
\begin{figure}
    \centering
    \includegraphics[trim={0.1cm 2cm 0 0.1cm},clip, totalheight=0.15\textheight]{\detokenize{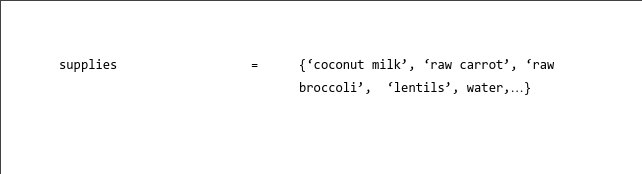}}
    \caption{An example of what $supplies$ might be.}
    \label{fig:$supplies$-example}
\end{figure}
With the addition of $mix\_dahl$ to the set $skills$\footnote{$kills$ is the set of all known processes, see chapter 2.}, ‘vegetable dahl’ can be fed into  SELECT_CONTENT (Algorithm \ref{alg:select-content}) which will perform the following steps. 
\begin{itemize}
\item check if ‘vegetable dahl’ is in $supplies$ and determine that it is not 
\item search $skills$ for a process which has ‘vegetable dahl’ in the  output, and find $mix\_dahl$, add $mix\_dahl$ to $action\_list$
\item check if the inputs to $mix\_dahl$, ‘fried vegetables’, ‘coconut  milk’ and ‘boiled lentils’ are in $supplies$ and determine that  ‘coconut milk’ is, add ‘coconut milk’ to $ingred\_list$ 
\item search for processes with ‘fried vegetables’ and ‘boiled lentils’ as outputs and find $fry\_vegetables$ and $strain\_lentils$ respectively,  add $fry\_vegetables$ and $strain\_lentils$ to $action\_list$ 
\item check if the inputs to $fry\_vegetables$ and $strain\_lentils$, ‘chopped  vegetables’ and ‘boiled lentils in boiling water’ are in  $supplies$ and determine that they are not 
\item search for processes with ‘chopped vegetables’ and ‘boiled  lentils in boiling water’ as outputs and find the ghost process  describing the synonym between ‘chopped broccoli’ and ‘chopped  carrot’ and ‘chopped vegetables’, and $boil\_lentils$, add ghost  and $boil\_lentils$ to $action\_list$ 
\item check if the inputs to $chop\_broccoli$, $chop\_carrot$ and $boil\_lentils$, ‘chopped  broccoli’, ‘chopped carrot’, ‘lentils’ and ‘boiling water’ are  in $supplies$, determine that ‘lentils’ is, add ‘lentils’ to  $ingred\_list$ 
\item search for processes with outputs ‘chopped broccoli’, ‘chopped  carrot’, and ‘boiling water’, and find $chop\_broccoli$,  $chop\_carrot$ and $bring\_to\_boil$ respectively, add $chop\_broccoli$,  $chop\_carrot$ and $bring\_to\_boil$ to $action\_list$ 
\item check if the inputs to $chop\_broccoli$, $$chop\_carrot$$ and  $bring\_to\_boil$, ‘raw broccoli’, ‘raw carrot’ and ‘water’, are in  $supplies$ and determine that they are, add ‘raw broccoli’, ‘raw  carrot’ and ‘water’ to $ingred\_list$
\item return $ingred\_list$ and $action\_list$ 
\end{itemize}

\begin{figure}
    \centering
    \includegraphics[trim={0.1cm 0 0 0.1cm},clip, totalheight=0.28\textheight]{\detokenize{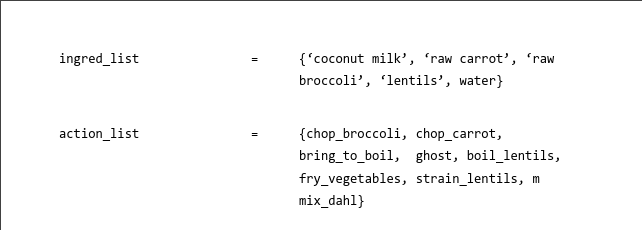}}
    \caption{Set of ingredients and actions after the content production phase.}
    \label{fig:formed-ingreds-and-actions}
\end{figure}

 Together $ingred\_list$ and $action\_list$ constitute the selected discourse content. This content is represented in Figure \ref{fig:formed-ingreds-and-actions}.
 
This discourse content allows the appropriate requires relations to be determined: $requires(A,B)$  whenever $A[\text{input}]$ contains an element from $B[\text{output}]$. 
\begin{figure}
    \centering
    \includegraphics[trim={0.1cm 2cm 0 2cm},clip, totalheight=0.53\textheight]{\detokenize{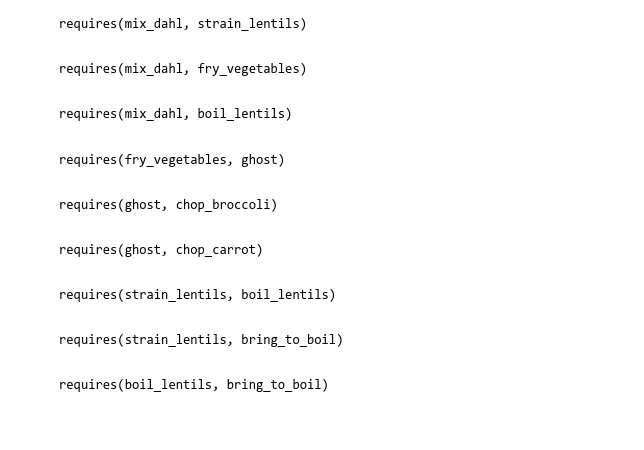}}
    \caption{All requires relations that hold between elements of $action\_list$.}
    \label{fig:requires-relations}
\end{figure}

\begin{figure}
    \centering
    \includegraphics[trim={0.1cm 0 0 0.1cm},clip, totalheight=0.95\textheight]{\detokenize{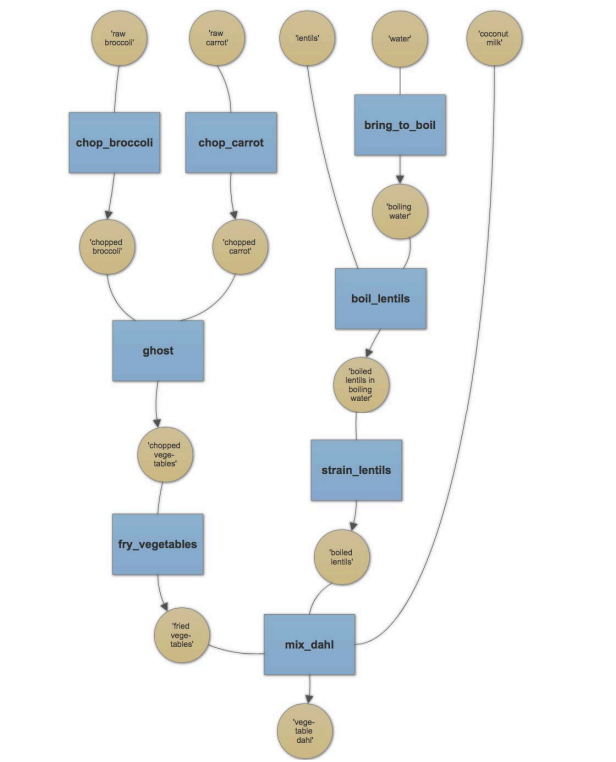}}
    \caption{Full DAG representing content of the generated recipe.}
    \label{fig:full-DAG}
\end{figure}

\begin{figure}
    \centering
    \includegraphics[trim={0.1cm 0 0 0.1cm},clip, totalheight=0.95\textheight]{\detokenize{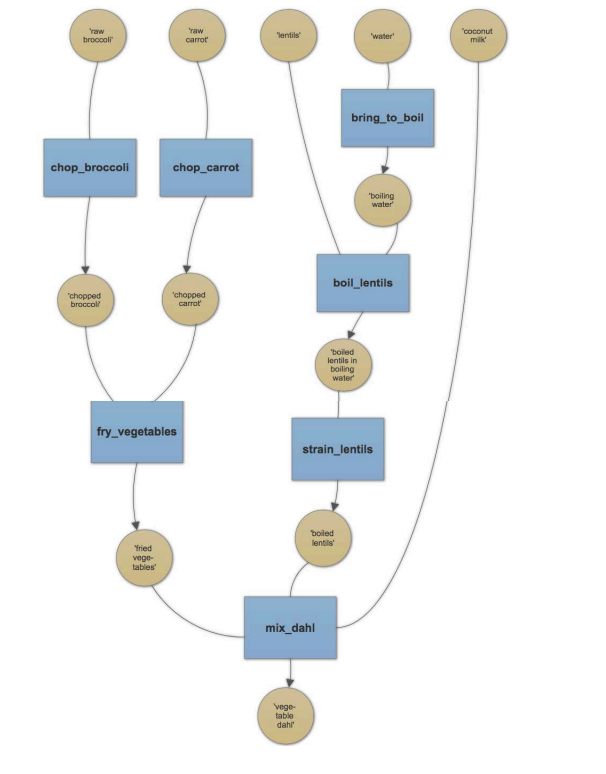}}
    \caption{DAG representing recipe content after the removal of ghost processes. Its inputs are joined to $fry\_vegetables$. The requires relations are updated as per Section \ref{subsec:find-all-lists}.}
    \label{fig:DAG-wo-ghosts}
\end{figure}

Descriptive strings are brown circles and processes are blue boxes (and arrows). Processes take a set of  descriptive strings as input and produce a set of descriptive strings as output. The figure is formed by  matching the inputs of some processes with the outputs of others. The five initial descriptive strings constitute $ingred\_list$, and the set of all processes constitutes $action\_list$.

\begin{figure}
    \centering
    \includegraphics[trim={0.1cm 0 0 0.1cm},clip, totalheight=0.18\textheight]{\detokenize{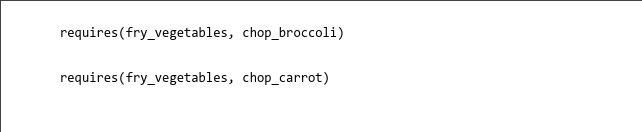}}
    \caption{New requires relations that are added after the removal of the ghost process.}
    \label{fig:new-requires}
\end{figure}

Next, the ghost is removed, giving Figure \ref{fig:DAG-wo-ghosts}, and $requires(A,B)$ is added whenever $requires(A,  ghost)$ and $requires(ghost, B)$ (see Section \ref{subsec:find-all-lists}), as shown in Figure \ref{fig:new-requires}. The final requires relations are depicted in Figure \ref{fig:requires-relations}. 

\begin{figure}
    \centering
    \includegraphics[trim={0.1cm 0 0 0.1cm},clip, totalheight=0.6\textheight]{\detokenize{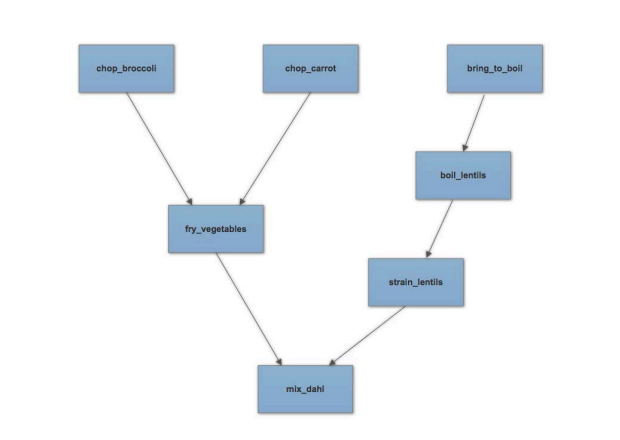}}
    \caption{DAG showing recipe content with processes only, no descriptive strings. An arrow from $P_1$ to $P_2$ indicates that requires(P2,P1). This form of representation can be termed a workflow and is discussed briefly  in Section \ref{subsec:current-lit}.}
    \label{fig:DAG-wo-strings}
\end{figure}

Then $action\_list$ is fed into FIND_ALL_LISTS, which determines all orderings of $action\_list$ such that no process precedes another which it requires. In this case 40 such orderings exist, so it is not feasible to describe the full action of the algorithm. Instead the first two steps are described and 3 of the output orders considered.
The first two stages to FIND_ALL_LISTS are shown below. 

\begin{figure}
    \centering
    \includegraphics[trim={2.1cm 0 0 0.1cm},clip, totalheight=0.45\textheight]{\detokenize{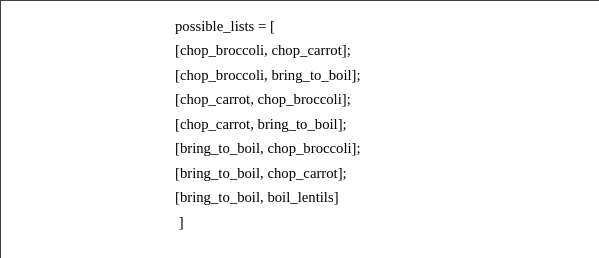}}
    \caption{All ordered pairs that can serve as prefixes to a permissible recipe for vegetable dahl.}
    \label{fig:two-elm-possible-lists}
\end{figure}

\begin{itemize}
\item search for the processes in $action\_list$ that do not require any others,  and determine these to be $chop\_broccoli$, $chop\_carrot$ and 
$bring\_to\_boil$, add these to the set $possible\_lists$ 
\item iterate through the set $possible\_lists$, for each element $P$ obtain all  elements that do not require any others except (possibly) $P$
\item for each of these elements $x$, add the list $[P,x]$ to possible lists, remove $P$ from  
$possible\_lists$ 
\item if $P$ = $chop\_broccoli$, then $x$ = $chop\_carrot$ or $bring\_to\_boil$;  giving the following two element lists: 
[$chop\_broccoli$, $chop\_carrot$] 
[$chop\_broccoli$, $bring\_to\_boil$] 
\item if $P$ = $chop\_carrot$, then $x$ = $chop\_broccoli$ or $bring\_to\_boil$; giving  the following two element lists: 
[$chop\_carrot$, $chop\_broccoli$] 
[$chop\_carrot$, $bring\_to\_boil$] 
\item if $P$ = $bring\_to\_boil$, then $x$ = $chop\_broccoli$ or $chop\_carrot$ or $boil\_lentils$; giving the following two element lists: 
[$bring\_to\_boil$, $chop\_broccoli$] 
[$bring\_to\_boil$, $chop\_carrot$] 
[$bring\_to\_boil$, $boil\_lentils$] 
\end{itemize}
At this point, $possible\_lists$ is composed of seven two-element lists, as shown in Figure \ref{fig:two-elm-possible-lists}.The next step would be to again iterate through $possible\_lists$ and, for each two-element list  $[P,Q]$, determine all the processes that do not require any others except for (possibly) $P$ and $Q$.  For example,  
\begin{itemize}
\item if $[P,Q]$ = [$chop\_broccoli$, $chop\_carrot$], then $x$ = $bring\_to\_boil$ or  $fry\_vegetables$; giving the following three element lists: 
[$chop\_broccoli$,$chop\_carrot$, $fry\_vegetables$]  
[$chop\_broccoli$, $chop\_carrot$, $bring\_to\_boil$]  
\end{itemize}

These three element lists would be added to $possible\_lists$, and $[chop\_carrot,  chop\_broccoli]$ would be removed.  
Of the 40 valid lists that would eventually be obtained by FIND_ALL_LISTS and hence fed into  the CONCURRENT_COMPRESSION algorithm, the following three are considered in this worked  example. The notation $P(x/y)$ refers to a process $P$ with total time $x$ and free time $y$. 
\ref{fig:two-elm-possible-lists}.
\begin{figure}
    \centering
    \includegraphics[trim={0.1cm 0 0 0.1cm},clip, totalheight=0.99\textheight]{\detokenize{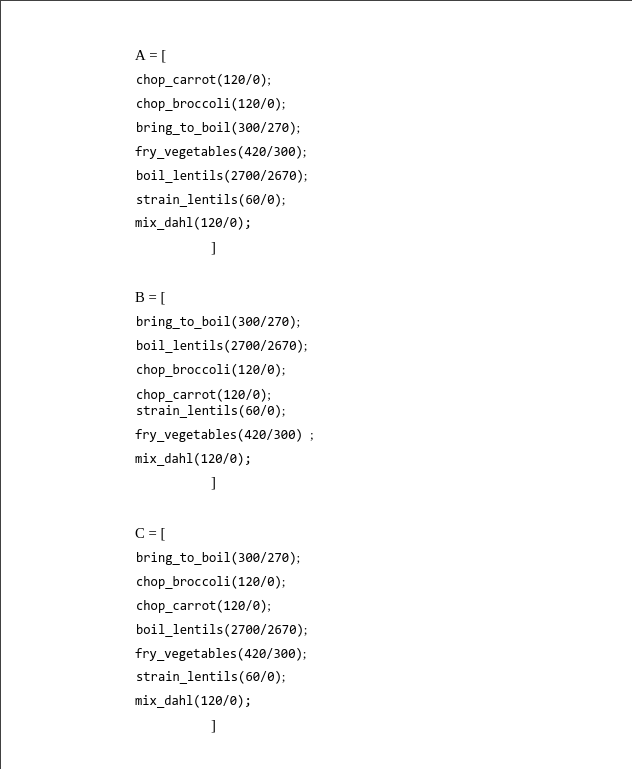}}
    \caption{Three of the permissible orders, out of a total of 40, for the steps in the recipe for vegetable dahl}
    \label{fig:ABC-recipe}
\end{figure}

CONCURRENT_COMPRESSION iterates through a list of processes and combines neighbouring  processes under certain conditions (see Algorithm \ref{alg:concurrent-compression} for a full description).  It’s action on $A$, $B$ and $C$ is described below. The notation $X'$ indicates the output of CONCURRENT_COMPRESSION after input $X$.  
\\

\textbf{Order A}
\begin{enumerate}
\item $mix\_dahl$ has no elements following it and so is passed over 
\item $strain\_lentils$ has f time = 0 and so is passed over 
\item $boil\_lentils$ has f time $> 0$, but is followed by $strain\_lentils$, and  requires($strain\_lentils$, $boil\_lentils$), so $boil\_lentils$ is passed  over 
\item $fry\_vegetables$ has f time $> 0$, but it is followed by $boil\_lentils$,  and $boil\_lentils[\text{time}] > fry\_vegetables[\text{f time}]$, so 
$fry\_vegetables$ is passed over 
\item $bring\_to\_boil$ has f time $ > 0$, but it has $fry\_vegetables$ to its right,  and $fry\_vegetables[\text{time}] > bring\_to\_boil[\text{f time}]$, so $bring\_to\_boil$ is passed over 
\item $chop\_broccoli$ has f time = 0 and so is passed over 
\item $chop\_carrot$ has f time = 0 and so is passed over 
\end{enumerate}
Thus, no processes are combined and the output from concurrent compression is the same as the input.  
A’ = A 
\\

\begin{figure}
    \centering
    \includegraphics[trim={0.1cm 1.4cm 0 0.1cm},clip, totalheight=0.22\textheight]{\detokenize{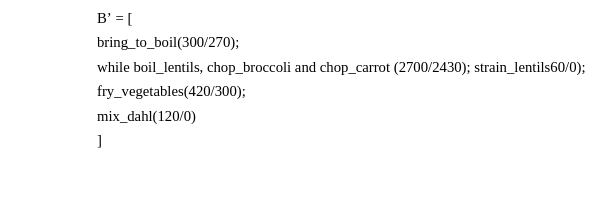}}
    \caption{Final, temporally optimized, recipe resulting from order B.}
    \label{fig:recipeB-compressed}
\end{figure}

\textbf{Order B}
\begin{enumerate}
\item $mix\_dahl$ has no elements following it and so is passed over 
\item $fry\_vegetables$ has f time $> 0$, but is followed by $mix\_dahl$ and  $requires(mix\_dahl, fry\_vegetables)$, so $fry\_vegetables$ is passed  over 
\item $strain\_lentils$, $chop\_carrot$ and $chop\_broccoli$ have f time = 0  and so are passed over 
\item $boil\_lentils$ has f time $> 0$, it is followed by $chop\_broccoli$,  $chop\_broccoli[\text{time}] < boil\_lentils[\text{f time}]$ and neither $requires(chop\_broccoli, boil\_lentils)$ nor $requires(boil\_lentils,  chop\_broccoli)$ hold; also, $boil\_lentils$ is not going to be inserted  into anything else in the future as the only process preceding it is $bring\_to\_boil$ and $requires(boil\_lentils, bring\_to\_boil)$; 
therefore $chop\_broccoli$ is inserted into $boil\_lentils$ and $boil\_lentils[\text{f time}]$ is then reduced by $chop\_broccoli[\text{time}]$ to  2670-120 = 2550
\item the next process after $boil\_lentils$ is $chop\_carrot$, which for the  same reasons is also inserted, $boil\_lentils[\text{f time}]$ is then reduced by $chop\_carrot[\text{time}]$ to 2550-120 = 2430 
\item the next process following $boil\_lentils$ is $strain\_lentils$ and  \\ $requires(strain\_lentils, boil\_lentils)$, so no more insertions into  $boil\_lentils$ can take place 
\item $bring\_to\_boil$ is followed by $boil\_lentils$, and $requires(boil\_lentils, bring\_to\_boil)$ so $bring\_to\_boil$ is passed  over 
\end{enumerate}
Thus, the only combination is the insertion of $chop\_broccoli$ and $chop\_carrot$ into $boil\_lentils$, and the output is as shown in Figure \ref{fig:recipeB-compressed}.
\ref{fig:recipeB-compressed}.
\\

\begin{figure}
    \centering
    \includegraphics[trim={0.1cm 1.3cm 0 0.1cm},clip, totalheight=0.22\textheight]{\detokenize{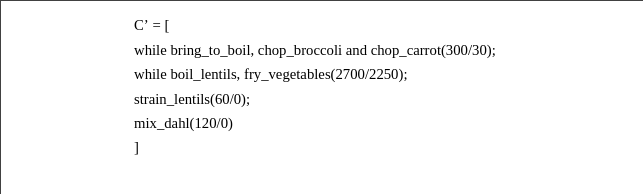}}
    \caption{Final, temporally optimized, recipe resulting from order C.}
    \label{fig:recipeC-compressed}
\end{figure}

\textbf{Order C}
\begin{enumerate}
\item $mix\_dahl$ has no elements following it and so is passed over 
\item $strain\_lentils$ has f time = 0 and so is passed over 
\item $fry\_vegetables$ has f time $> 0$, it is followed by $strain\_lentils$, \\
$strain\_lentils[\text{time}] < fry\_vegetables[\text{f time}]$ and \\ $requires(strain\_lentils, fry\_vegetables)$ does not hold; however,  
$fry\_vegetables$ is preceded by $boil\_lentils$ and the conditions for combination hold between $fry\_vegetables$ and $boil\_lentils$,  moreover the insertion of $strain\_lentils$ into $fry\_vegetables$ would prevent the insertion of $fry\_vegetables$ into $boil\_lentils$ as  $requires(strain\_lentils, boil\_lentils)$ and so it would be  infelicitous to have while boil lentils, (while fry vegetables,  strain lentils); therefore the insertion of $strain\_lentils$ into  $fry\_vegetables$ is foregone in favour of the future insertion of  $fry\_vegetables$ into $boil\_lentils$ 
\item $boil\_lentils$ is followed by $fry\_vegetables$, \\ $boil\_lentils[f  time] > fry\_vegetables[\text{time}]$ and $requires(boil\_lentils,fry\_vegetables)$ does not hold, moreover  $boil\_lentils$ is not going to be inserted into anything else in the future  as $chop\_broccoli[\text{f time}]= chop\_carrot[\text{f time}] = 0$ and requires ($boil\_lentils,bring\_to\_boil$); therefore $fry\_vegetables$ is  inserted into $boil\_lentils$, $boil\_lentils$[\text{f time}] is then reduced  by $fry\_vegetables$[\text{time}] to give 2670-420 = 2250 the next element  following $boil\_lentils$ is $strain\_lentils$ and  
requires($strain\_lentils$, $boil\_lentils$), therefore no more  insertions into $boil\_lentils$ can take place 
\item $chop\_broccoli$ and $chop\_carrot$ have f time = 0 and so are passed  over 
\item $bring\_to\_boil$ is followed by $chop\_broccoli$, $bring\_to\_boil$[f  time] > $chop\_broccoli$[\text{time}] and requires($bring\_to\_boil$,  $chop\_broccoli$) does not hold,  moreover $bring\_to\_boil$ is not going to be combined with anything  else in the future as it has no elements preceding it; therefore  $chop\_broccoli$ is inserted into $bring\_to\_boil$; $bring\_to\_boil[\text{f time}]$ is then reduced by $chop\_broccoli[\text{time}]$ to 270 – 120 = 150 the  next process following $bring\_to\_boil$ is $chop\_carrot$, and for the  same reasons it too is inserted into $bring\_to\_boil$, $bring\_to\_boil[\text{f time}]$ is then reduced by $chop\_carrot[\text{time}]$ to 150-120 = 30
\item the next process following $bring\_to\_boil$ is $boil\_lentils$ and \\ $requires(boil\_lentils,bring\_to\_boil)$, therefore nothing else is inserted into $bring\_to\_boil$ 
\end{enumerate}
Thus $fry\_vegetables$ was inserted into $boil\_lentils$, and $chop\_broccoli$ and $chop\_carrot$ were inserted into $bring\_to\_boil$, and the output is as shown in Figure \ref{fig:recipeC-compressed}.

\begin{figure}
    \centering
    \includegraphics[trim={0.1cm 0 0 0.1cm},clip, totalheight=0.75
    \textheight]{\detokenize{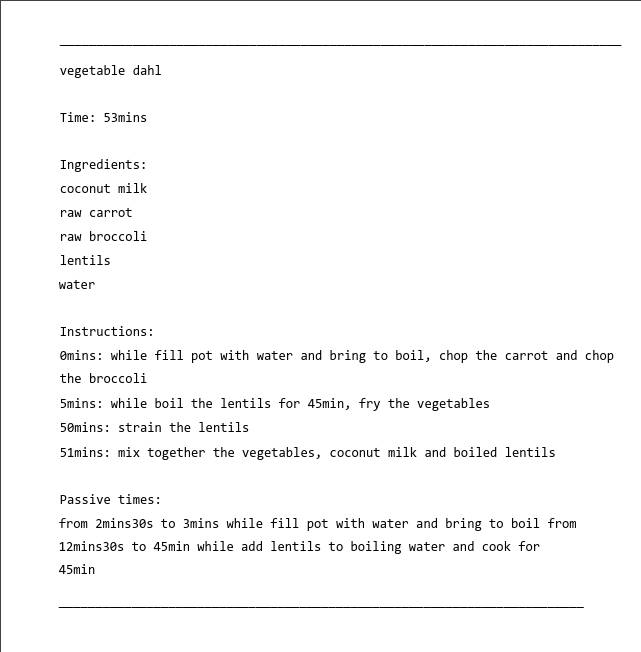}}
    \caption{The recipe generated by FASTFOOD in response to a user input of ‘vegetable dahl’.}
    \label{fig:veg-dahl-recipe2}
\end{figure}

The total times for the compressed recipe plans $A'$,$B'$ and $C'$ are 
\begin{align*}
T_{A'} &= 120 + 120 + 300 + 420 + 2700 + 60 + 120 = \boldsymbol{3840} \\
T_{B'} &= 300 + 2700 + 60 + 420 + 120 = \boldsymbol{3600} \\
T_{C'} &= 300 + 2700 + 60 + 120 = \boldsymbol{3180} \,,
\end{align*}
where $T_x$ denotes the total time for recipe plan $x$. Thus, $C$ is the most efficient as it contains the largest amount of combinations. In fact of the 40 permissible orderings of these processes, no other, when fed through CONCURRENT_COMPRESSION, has total time less than that of $C'$. Thus,  $C'$ is chosen as the recipe plan.  
The final requirement is the temporal information: the total cooking time and the passive times. This can be obtained by iterating through each element in $C'$ and noting those that still have  strictly positive free time (remembering that free time will have been reduced in instances of  insertion). The variable $$clock$$ is used to keep track of when these free times will occur. The notation ‘while $P$, $P_1, \dots, P_n$’ refers to the  combination that results from inserting $P_1, \dots, P_n$ into $P$.
\begin{itemize}
\item $clock$ is set to 0 
\item $passives$ is set to {} 
\item $clock$ is increased by the time value of while $bring\_to\_boil$,  $chop\_broccoli$ and $chop\_carrot$, giving 0 + 300 = 300
\item while $bring\_to\_boil$, $chop\_broccoli$ and $chop\_carrot$ has f  time = 30 so the triple (270,300, `while fill pot with water and bring to boil’) is marked as passive; the instruction of this process is also noted, so 
((270,300),`fill pot with water and bring to boil’) is added to $passives$ 
\item $clock$ is increased by the time value of while $boil\_lentils$,  $fry\_vegetables$ , giving 300 + 2700 = 3000 
\item while $boil\_lentils$, $fry\_vegetables$ has f time = 2250 so the  triple (750,3000, `while boil the lentils’)\footnote{The original direction of $boil\_lentils$ is stored on insertion, so that it can be used in $passives$.} is marked as passive; the instruction of this process is also noted, so the triple (750,  
3000, `while place lentils in boiling water and cook for 45min, stirring occasionally’) is added to $passives$ 
\item $clock$ is increased by $strain\_lentils$[\text{time}], giving 3000 + 60 = 3060 
\item $strain\_lentils$ has no passive time 
\item $clock$ is increased by $mix\_dahl[\text{time}]$, giving 3060 + 120 = 3180 
\item $mix\_dahl$ has no passive time 
\item $clock$ is saved as the total cooking time 
\end{itemize}
Now the recipe can be generated from the set $ingred\_list$, the recipe plan $C'$ and the temporal  information. The ingredients list is a printing of every element of $ingred\_list$; the list of  instructions is a printing of the instruction of each element in $C'$ along with the time at which  they are to be performed (this is obtained using a $clock$ variable as in $passives$); and the  description of passive times is a printing of the information in each triple $x$ in $passives$ in the  format: ‘from $x[0]$ to $x[1]$ while $x[2]$’. The total time for the recipe is also printed. The final generated recipe for ‘vegetable dahl’ is shown in Figure \ref{fig:veg-dahl-recipe2}. 
 
This example demonstrates that the representation system and NLG methods discussed in this  paper are capable of generating a cooking discourse which is understandable, albeit slightly  unnatural. It also shows that the temporal optimisation procedure (Section \ref{sec:content-organization}) is effective, clearly there are time savings by performing the instructions concurrently where directed, rather than one at a time. It is claimed that the temporal optimisation algorithm  will always obtain an optimal solution, which here implies there is no faster way of performing  the steps in this recipe than the method indicated; that, based on the time assumed for each step,  the recipe cannot be executed in less than 53 minutes. However, a proof of the guarantee of  optimality is outside the scope of the present paper. 
More practically, there are some advantages and some disadvantages to this generated recipe as  compared with a recipe from a cookbook.  
\\

\textbf{ADVANTAGES}
\begin{itemize}
\item The language is uniform and concise. Something which has been identified as  desirable in cooking recipes \cite{mori2012machine} and advantageous to users in other practical domains \cite{reiter2005choosing}.
\item It allowed the user to input the foods on hand and then receive a yes or no as to  whether the dish was makeable (yes in this case). This is perhaps an easier task  than the corresponding task of the user of a cookbook: for every item in the  ingredient list, checking whether it is on hand.  
\item It gives a precise indication of the time taken, although this is based on  
assumed times for each step and so may vary depending on the cook.  
\item It saves the cook the effort of deciding which steps to perform simultaneously,  as this is indicated explicitly in every case.  
\item It displays the times at which the cook will be passive. Two recipes may have  the same total time, but differ significantly in how much free time they include  and consequently in how demanding they are. In this case the cook will be free  for over 30min at a stretch (again based on assumed times for each step), and  this is useful information to know at a glance.  
\end{itemize}

\textbf{DISADVANTAGES}
\begin{itemize}
\item There is no notion of quantity, no specification of the amount of each  ingredient required. It is thought that an extension to include quantity would  not be overly difficult within the existing framework, though the function of  FASTFOOD is currently impoverished in the absence of such an extension.  
\item The language is less natural than would be desired. A particular problem is the  lack of inflection for progressive aspect in the combined instructions:  
‘while boil the lentils..’ instead of ‘while boiling the lentils..’.  
\item A more complete version would have considered alternative ways of making  the dish, e.g. by using different vegetables. 
\end{itemize}
The presentation of this recipe for vegetable dahl is partially an academic point in support of the  NLG techniques discussed in this paper. However it also presents some practical advantages  over cookbook recipes. These advantages may become more significant if the disadvantages  were mitigated through various extensions and improvements to the program, most notably the  inclusion of the notion of quantity and the correct inflection of certain verbs in the output text. 

%% file: related_work.tex
\section{Related Work} \label{sec:related-work}
The majority of recent computational approaches to cooking recipes have focussed either on  searching and annotating large online databases, or on deriving a graphical depiction of a given  recipe in terms of a workflow. These are enterprises  of Natural Language Understanding (NLU), which is a fundamentally different approach to the  recipe generation attempted by FASTFOOD. However, there exists an early and detailed system  which also takes the generative approach in Dale’s EPICURE (1992). In Section \ref{subsec:epicure-comparison} a comparison  is made between FASTFOOD and EPICURE, and it is argued that, though they are both concerned  with recipe generation as opposed to recipe understanding, there is a clear difference in the  strengths and weaknesses of the two systems. Section \ref{subsec:current-lit} some current NLU work on cooking  recipes. An overview is given of a technique for extracting workflow representation, along with  a problem this technique can face. It is shown that, by contrast, a workflow representation is readily achieved  by FASTFOOD.

\subsection{Comparison with EPICURE} \label{subsec:epicure-comparison}
EPICURE is a program for recipe generation proposed by Dale (\citeyear{dale1989cooking}, \citeyear{dale1990generating}). It  contains a representation system for recipe discourse and a method of generating a text  describing this recipe. It is similar to FASTFOOD in that its focus is on NLG as opposed to NLU,  and the difficulties identified by Dale (1992) in semantically representing cooking discourse are  similar to those identified in Section \ref{subsec:discourse-ref-problems} of the present paper. Thus, EPICURE and FASTFOOD both aim to achieve the same goal—a method of text generation in a domain with mutable  entities, see Section \ref{subsec:discourse-ref-problems} and Dale \citeyear{dale1991content}, Chapter 2. 

However, there are considerable differences  in the structure and performance of the two programs.  As discussed in the introduction, NLG systems can generally be thought of as first performing  strategic choices of ‘what to say’, and then moving towards a tactical specification of ‘how to  say it’ (McKweon, 1992, 1985). This framework can provide a fruitful comparison between  FASTFOOD and EPICURE. Within such a comparison FASTFOOD focusses primarily on the ‘what  to say’ component of NLG. Two of its primary goals are to employ an ATP style algorithm to  determine the steps necessary in a cooking recipe (Section \ref{sec:content-selection}), and to derive a time efficient way  of performing these steps (Section \ref{sec:content-organization}), with comparatively less emphasis on the ‘how to say it’ component: the output text  is a sequence of imperatives constructed from a representation of cooking actions such as ‘chop’  and food classes such as ‘carrot’ (see Section \ref{sec:content-realization}). There is no scope for pronominal reduction—
it may for example contain a fragment such as ‘peel the carrot, chop the carrot, roast the carrot’,  instead of ‘peel chop and roast the carrot’—and in certain contexts verbs do not receive the  correct inflection—‘while boil the lentils..’ instead of ‘while boiling the lentils…’. EPICURE, on  the other hand, is at least as concerned with tactical realisation as with strategic planning. Two primary goals of the program are to correctly generate anaphora (both pronouns and one anaphora) and to determine when separate ingredients should be listed together e.g. ‘salt and  pepper’. 

The most striking manifestation of this difference in emphasis is in the treatment of discourse  referents. As discussed in Section \ref{subsec:discourse-ref-problems}, these are not used by FASTFOOD at all, whereas they are  employed in a detailed manner by EPICURE. Indeed, the system is part of a series of works by  the author on the generation of referring expressions \cite{dale1989cooking, dale1991content, dale2007handling, viethen2007evaluation}. The semantics of the discourse in EPICURE is modelled  using these discourse referents and a set of states corresponding to time points. 
\begin{quotation}
the current state of each entity is modelled by a knowledge base object  
representing all properties that are true of that entity in that state; the world  model is just that collection of such objects that describes all the entities in the  domain. Whenever an action takes place, this collection of objects is  
updated... (Dale, 1992, p71.).  
\end{quotation}

Referring expressions are formed from discourse referents, and a discourse referent is a model  of an entity’s function in the discourse. This means that a referring expression can be informed  by the function of the entity it refers to. A discourse fragment in which the same entity performs  first one function and then another can be realised by one clause rather than two—’chop and  peel the carrot’--and similarly for a fragment in which two different entities have the same  function-- ‘chop the carrot and broccoli’. This is an advantage of choosing a referent-based  discourse representation.  

However, there are also disadvantages to the representation system of EPICURE as compared  with FASTFOOD. Firstly, it is less efficient in its content generation, deriving its discourse content, ‘what to say’,  from a plan library that, for a given dish, explicitly contains the steps necessary to make it. For example, the fact that beans can be prepared by first soaking, then draining and then rinsing  is hard-coded into the program. FASTFOOD on the other hand, creates its discourse  content by selecting elements from a database of descriptive strings and processes, (Section \ref{sec:content-selection})  with this database in turn being produced by more fundamental structures corresponding to  foods such as carrot, a cooking actions such as chopping (Section \ref{sec:content-production}). Being able to create discourse content is preferable to requiring a ready-made version as it demonstrates an extra level  of intelligence to the program and reduces the need for manual specification. The parts of recipes—steps and ingredients—can be combined in numerous ways; there exist many  more recipes than parts of recipes. Consequently it is more efficient to encode only the latter (in this case  processes and descriptive strings), and allow the recipes to be created by the program through various  combinations of these parts. Even greater efficiency is obtained in FASTFOOD by, in turn, producing  these parts from more abstract structures (Section \ref{sec:content-production}).

There are strengths and weaknesses to both systems. EPICURE’s more detailed discourse  representation, in terms of discourse referents and states holding at a sequence of time points,  produces a smoother and more natural output text but requires pre-existing specification of  discourse content. FASTFOOD on the other hand, uses a lighter representation of discourse as a  set of processes and descriptive strings. This sometimes produces unnatural output text but  means that discourse content can be formed in response to user input by selecting the appropriate  elements from a database of processes and descriptive strings. An ideal solution would be a combination of the two methods which maintained the advantages of each, and this could  possibly be achieved by describing the processes in FASTFOOD as transitions between the state  descriptions in EPICURE. Currently, processes are represented as transitions between units  termed descriptive strings whose only function is to semantically represent states of affairs (see  Section \ref{sec:representation-system}.2), but in theory descriptive strings could be replaced with anything else that serves  the same function. The state descriptions in EPICURE are relatively complex, they are sets of  multiple objects each described in detail, but the processes of FASTFOOD could potentially be  redefined based on an input state and output state, with these states in turn defined as sets of  EPICURE objects predicated of by cooking-related properties such as ‘soaked’ and ‘whole’, as  well as grammatical properties such as ‘count noun’ and ‘plural’. This may allow the recipe to  be generated and optimised by an architecture like that of FASTFOOD, with the resulting  discourse being represented by the discourse referents and states of EPICURE, thereby producing  the best of both worlds.  
 
EPICURE is, in a broad sense, similar to FASTFOOD. Both systems attempt recipe generation from a pre-existing set of internal structures. However, where FASTFOOD focusses to a greater extent  on the strategic choice of what to say, EPICURE shows comparatively greater concern for  tactically specifying how to say it. The main result of this difference is that, compared with  FASTFOOD, EPICURE produces smoother and more natural discourse, but requires considerably  more detailed pre-existing structures, entire recipe plans rather than foods and cooking actions, and cannot perform abstract reasoning on the cooking discourse content.  A broad sketch was given of a system that combines the two representation methods, potentially  offering the advantages of both programs.

\subsection{Comparison with Current Literature} \label{subsec:current-lit}
FASTFOOD, similar to EPICURE \cite{dale1989cooking} discussed in Section \ref{subsec:epicure-comparison}, is a program for Natural  Language Generation. Most recent work on cooking recipes however, approaches the problem  from the side of Natural Language Understanding (NLU), with the goal of analysing a natural  language recipe and extracting a more abstract representation.  
The purpose of such extraction can be quite varied: recipe search \cite{yamakata2017cooking}, recipe recommendation and alteration \cite{gunamgari2014hierarchical, cordier2014taaable, gaillard2012extracting}; recipe  comparison \cite{mori2012machine}; instruction of robotic cooking  assistants (\cite{zhaotraining}; or translation recipes from one natural language to another \cite{mori2014flow}. However the representation systems themselves are more uniform.  While some technical systems exist for annotating recipes with logical or computer language  notation \cite{cazzulino2004beginning}, the majority of recent work has used  the intuitive method of representing recipes as workflows. The technical structure for such a  representation is a Directed Acyclic Graph (DAG) which is composed of vertices (cooking  steps) some of which are joined by arrows (indicating which steps depend on, and must be  performed before, which others). An example of a workflow representation can be seen in Figure \ref{fig:DAG-wo-strings}, which represents FASTFOOD’s recipe for vegetable dahl.  

Expressing a cooking recipe in the form of a workflow requires identifying the steps in the recipe  and the dependencies that exist between them. The typical method for extracting workflows from procedural texts has been described in general by \cite{abend2015lexical}, and  in the specific domain of cooking recipes by \cite{mori2014flow, dufour2012semi, hamada2000structural}. It uses a combination of Named Entity Recognition (NER) and Predicate Argument  Analysis (PAA). Firstly, NER is performed on each instruction to extract the name\footnote{See Section \ref{subsec:things-vs-descriptions} of the present work for a brief philosophical discussion of the ontological properties of  names in cooking recipes.} of the food  being operated on. Next, a predicate-argument structure is formed for each verb, essentially by  pairing it with the nearest noun that was identified in NER. Formally, these extracted entities  correspond to the vertex set of the DAG. Finally the dependencies between these structures— formally the edges of the DAG—are determined using the order in which they appeared in the  text—a predicate argument structure $A$ depends on another $B$ if they share the same argument and $A$ appears later in the text than $B$. The recipe fragment ‘peel the carrot, chop the carrot’  would be represented as ‘peel(carrot), chop(carrot)’ and so the latter would be said to depend on  the former, i.e. they would be joined with a directed edge.  
However, this technique faces a problem in certain circumstances which comes back to the  transitory quality of the ontology of a cooking discourse (see Section \ref{subsec:discourse-ref-problems}). Specifically it would  seem unable to identify dependencies in cases where new entities have been introduced.  Consider (7.1), an online recipe for sweet potato burgers. 
\begin{verse} \label{burger-patty-problem}
1. chop cilantro and leaves \\
2. bake patties for 35 min \\
3. place burgers on burger bun
\end{verse} 
The entity ‘burgers’ has never been mentioned before, so there would be no guide as to which  nodes it should be joined, no reason for 3 to be joined to 2. If it were specified that every node  must be joined to at least one preceding node (formally that the DAG be connected) then there  would be nothing in the above technique to indicate that 3 should be joined to 2 rather than 1. 

Of course it is possible to supplement this technique in an attempt to address such problems. For  example \cite{hamada2000structural} also employ a dictionary of cooking items. However, even a  dictionary may fail to resolve this problem: the entry for ‘burgers’ does not contain the word ‘patties’ in  any of 4 large online dictionaries (Collins Dictionary 2017; Oxford Living Dictionary 2017;  Cambridge Advanced Learner’s Dictionary and Thesaurus, 2017; Merriam-Webster Online  2017), so it is not clear how this could directly provide the missing link between 3 and 2. Thus  the common methods of extracting workflow representations of cooking recipes can face  problems in certain circumstances, such as the above example.  

In FASTFOOD on the other hand, workflow representation is readily attained by the methods  described in Sections \ref{sec:content-production} and \ref{sec:content-selection}. Indeed it is this representation, including its dependencies, that is  ultimately used to generate the text. Figure \ref{fig:DAG-wo-strings} is a graphical depiction of this representation in  the case of the worked example of Section \ref{sec:worked-example}. This can be seen as a more general advantage of  exploring the NLG of cooking recipes: when the representational elements are created by the  system itself, they can be created in a way that facilitates the extraction of whatever information  is most pertinent. In this case, representing cooking instructions in terms of input and output  states (Section \ref{sec:representation-system}) allows dependencies to be readily established by matching the input of one  with the output of another (Section \ref{subsec:optimization-problem-overview}).

These dependency relations were used in FASTFOOD’s temporal optimisation module (Section \ref{sec:content-organization}). Such optimisation required the consideration of multiple orders for the steps in the cooking  recipe, and accurate identification of dependency relations was necessary to establish which  orders produces coherent recipes (see Section \ref{subsec:optimization-problem-overview}). Further to this point, the only account of  recipe optimisation in the surveyed literature \cite{hamada2005cooking}, observed that automated DAG extraction was insufficiently accurate for their purposes,  and that a manual annotation was required instead.\footnote{The optimization attempted by \cite{hamada2005cooking} is different to that of  FASTFOOD as it was not solely concerned with temporal optimisation, but also the optimisation of other  factors such as utensil use.} Thus it can also be said that the NLG  approach to cooking recipes is helpful to their temporal optimisation: generating (as opposed to  extracting) discourse representations aids in the accurate identification of dependency relations,  and this in turn is necessary for any optimisation process that considers reorderings.

%% file: conclusion.tex
\section{Conclusion} \label{sec:conclusion} 
This paper introduced FASTFOOD an NLG system that use a  backwards-chaining Automated Theorem Prover (ATP) to select the correct steps  and ingredients required to make a given dish, and then modified the recipe to optimize time efficiency in the generated recipes. 
NLG in FASTFOOD was analysed as operating over four distinct phases:  content production, content selection, content organisation and content realisation, and these were described in detail in Sections \ref{sec:content-production}, \ref{sec:content-selection}, \ref{sec:content-organization} and \ref{sec:content-realization}, respectively. The use of backwards-chaining ATP to generate cooking recipes is of theoretical interest, as it draws a connection between two ostensibly disparate areas; and also of practical use, as it allows the NLG process to be modularised so that a database of representational elements is produced and stored, and  then elements from this database are selected in response to various user inputs.
Temporal optimisation shows the value of the representation system used in FASTFOOD, i.e. that its representation system facilitates this type of high-level reasoning. It also presents a practical benefit, taking care of a reasoning task normally left to the human cook, and reducing cooking time by determining the most time-efficient way to execute a recipe. 
A number of theoretical points of interest were also encountered in the design choices of FASTFOOD. Choosing what was believed to be the most suitable representation system (Section \ref{sec:representation-system}) meant  omitting discourse referents; and this decision displayed a connection to an established  philosophical debate on the nature of names and descriptions.  The modularistion resulting from the content selection algorithm, meant that a method was  required for producing a database of representational elements (Section \ref{sec:content-production}). Part of this  productive method employed a structure which touched on various semantic interpretations of  the synonymy relation (Section \ref{subsec:synonyms-and-ghosts}). The content selection algorithm, while initially inspired by ATP, also exhibited a similarity to  automated parsing, and this led to a three-way comparison between ATP, automated parsing and  FASTFOOD’s content selection algorithm. 
Finally, this work was compared to existing literature on Natural Language Processing of  cooking recipes (Section \ref{sec:related-work}, especially EPICURE\cite{dale1991content}. The two were shown to have different strengths  and weaknesses, and a way to combine the benefits of both was briefly explored. A comparison was also made with a common NLU technique of converting natural language recipes to directed acyclic graphs. It was shown that, due in part to its representation system  (Section \ref{sec:representation-system}), FASTFOOD avoids a common problem encounted by such techniques.
Future work includes introducing a notion of quantity, allowing the user to choose between multiple recipes for the same dish, and linking with the field of text image processing.